%% file: main.tex
\documentclass[10pt]{article} 
\usepackage[accepted]{tmlr}
\input{math_commands.tex}

\usepackage{microtype}
\usepackage{graphicx}
\usepackage{subfigure}
\usepackage{booktabs} 
\usepackage{hyperref}

\usepackage{amsmath}
\usepackage{amssymb}
\usepackage{mathtools}
\usepackage{amsthm}
\usepackage[capitalize,noabbrev]{cleveref}

\usepackage[font=small,skip=0pt]{caption}
\usepackage{appendix}
\usepackage{algorithm}
\usepackage{algpseudocode}

\usepackage{url}
\usepackage{mathrsfs}
\usepackage{verbatim}
\usepackage{graphicx}
\usepackage{amsfonts,amssymb,amsmath}
\usepackage{bm}
\usepackage{makecell,multirow}
\usepackage{eso-pic}
\usepackage{float}
\usepackage{subfigure}
\usepackage{booktabs}  
\usepackage{adjustbox}
\usepackage{bbm}
\usepackage{epstopdf}
\usepackage{threeparttable}
\usepackage{tablefootnote}
\usepackage{enumitem}
\usepackage{tikz}
\usepackage{multicol}
\usepackage{soul}

\theoremstyle{plain}

\theoremstyle{definition}

\theoremstyle{remark}

\usepackage[textsize=tiny]{todonotes}

\title{DHA: End-to-End Joint Optimization of Data Augmentation Policy, Hyper-parameter and Architecture}

\author{\name Kaichen Zhou  \\
      \addr University of Oxford
      \AND
      \name Lanqing Hong$^\dagger$ \email honglanqing@huawei.com \\
      \addr Huawei Noah's Ark Lab
      \AND
      \name Shoukang Hu \\
      \addr The Chinese University of Hong Kong
      \AND
      \name Fengwei Zhou \\
      \addr Huawei Noah's Ark Lab
      \AND
      \name Binxin Ru \\
      \addr SailYond Technology $\&$ Research Institute of Tsinghua University in Shenzhen
      \AND
      \name Jiashi Feng  \\
      \addr National University of Singapore
      \AND
      \name Zhenguo Li \\
      \addr Huawei Noah's Ark Lab}

\def\month{10}  
\def\year{2022} 
\def\openreview{\url{https://openreview.net/forum?id=MHOAEiTlen}} 

\begin{document}

\maketitle

\begin{abstract}
Automated machine learning (AutoML) usually involves several crucial components, such as Data Augmentation (DA) policy, Hyper-Parameter Optimization (HPO), and Neural Architecture Search (NAS). 
Although many strategies have been developed for automating these components in separation, joint optimization of these components remains challenging due to the largely increased search dimension and the variant input types of each component. In parallel to this, the common practice of \textit{searching} for the optimal architecture first and then \textit{retraining} it before deployment in NAS often suffers from low performance correlation between the searching and retraining stages. An end-to-end solution that integrates the AutoML components and returns a ready-to-use model at the end of the search is desirable.
In view of these, we propose \textbf{DHA}, which achieves joint optimization of \textbf{D}ata augmentation policy, \textbf{H}yper-parameter and \textbf{A}rchitecture. Specifically, end-to-end NAS is achieved in a differentiable manner by optimizing a compressed lower-dimensional feature space, while DA policy and HPO are regarded as dynamic schedulers, which adapt themselves to the update of network parameters and network architecture at the same time. Experiments show that DHA achieves state-of-the-art (SOTA) results on various datasets and search spaces. 
To the best of our knowledge, we are the first to efficiently and jointly optimize DA policy, NAS, and HPO in an end-to-end manner without retraining. 
\end{abstract}

\section{Introduction}
While deep learning has achieved remarkable progress in various tasks such as computer vision and natural language processing, the design and training of a well-performing deep neural architecture for a specific task usually requires tremendous human involvement \citep{he2016deep,sandler2018mobilenetv2}.
To alleviate such burden on human users, AutoML algorithms have been proposed in recent years to automate the pipeline of designing and training a model, such as automated Data Augmentation (DA), Hyper-Parameter Optimization (HPO), and Neural Architecture Search (NAS) \citep{cubuk2018autoaugment,mittal2020hyperstar,chen2019progressive}.
All of these AutoML components are normally processed independently and the naive solution of applying them sequentially in separate stages,
not only suffers from low efficiency but also leads to sub-optimal results \citep{dai2020fbnetv3,dong2020autohas}.
Indeed, how to achieve full-pipeline “from data to model” automation efficiently and effectively is still a challenging and open problem.

One of the main difficulties lies in understanding how to automatically combine the different AutoML components (e.g., NAS and HPO) appropriately without human expertise. 
FBNetV3~\citep{dai2020fbnetv3} and AutoHAS~\citep{dong2020autohas} investigated the joint optimization of NAS and HPO, while \citep{DBLP:journals/corr/abs-2012-09407} focused on the joint optimization of neural architectures and data augmentation policies. The joint optimization of NAS and quantization policy were also investigated in APQ~\citep{wang2020apq}. Clear benefits can be seen in the above works when optimizing two AutoML components together, which motivates the further investigation of ``from data to model" automation. However, with the increasing number of AutoML components, the search space complexity is increased by several orders of magnitudes and it is challenging to operate in such a large search space. In addition, how these AutoML components affect each other when optimized together is still unclear. Thus, further investigation is needed to open the black box of optimizing different AutoML components jointly.

Another main challenge of achieving the automated pipeline ``from data to model" is understanding how to perform end-to-end searching and training of models without the need of parameter retraining. 
Current approaches, even those considering only one AutoML component such as NAS algorithms, usually require two stages, one for searching and one for retraining~\citep{liu2018darts,xie2018snas}.
Similarly, automatic DA methods such as FastAA~\citep{lim2019fast} also need to retrain the model parameters once the DA policies have been searched.
In these cases, whether the searched architectures or DA policies would perform well after retraining is questionable, due to the inevitable difference of training setup between the searching and retraining stages \citep{yang2019evaluation}. To improve the performance correlation between searching and retraining stages, DSNAS~\citep{hu2020dsnas} developed a differentiable NAS method to provide direct NAS without parameter retraining. OnlineAugment~\citep{tang2020onlineaugment} and OnlineHPO~\citep{im2021online} design direct DA or HPO policy, respectively, without model retraining.

Targeting the challenging task-specific end-to-end AutoML, we propose DHA, a differentiable joint optimization solution for efficient end-to-end AutoML components, including DA, HPO and NAS. 
Specifically, the DA and HPO are regarded as dynamic schedulers, which adapt themselves to the update of network parameters and network architecture. 
At the same time, the end-to-end NAS optimization is realized in a differentiable manner with the help of sparse coding method, 
which means that instead of performing our search in a high-dimensional network architecture space, we optimize a compressed lower-dimensional feature space. With this differentiable manner, DHA can effectively deal with the huge search space and the high optimization complexity caused by the joint optimization problem. To summarize, our main contributions are as follows:
\begin{itemize}
    \item We propose an AutoML method, DHA, for the concurrent optimization of DA, HPO, and NAS. To the best of our knowledge, we are the first to efficiently and jointly realize DA, HPO, and NAS in an end-to-end manner without retraining. 

    \item Experiments show that DHA achieves a state-of-the-art (SOTA) accuracy on ImageNet with both cell-based and Mobilenet-like architecture search space. DHA also provides SOTA results on various datasets, including CIFAR10, CIFAR100, SPORT8, MIT67, FLOWERS102 and ImageNet with relatively low computational cost, showing the effectiveness and efficiency of joint optimization (see Fig.~\ref{fig:performance comparison}).
    
    \item Through extensive experiments, we demonstrate the advantages of doing joint-training over optimizing each AutoML component in sequence. Besides, higher model performance and a smoother loss landscape are achieved by our proposed DHA method.
\end{itemize}

\begin{figure*}[t]
\vspace{-0.3cm}
\setlength{\belowcaptionskip}{-0.5cm}
    \centering
    \includegraphics[width=17.2cm]{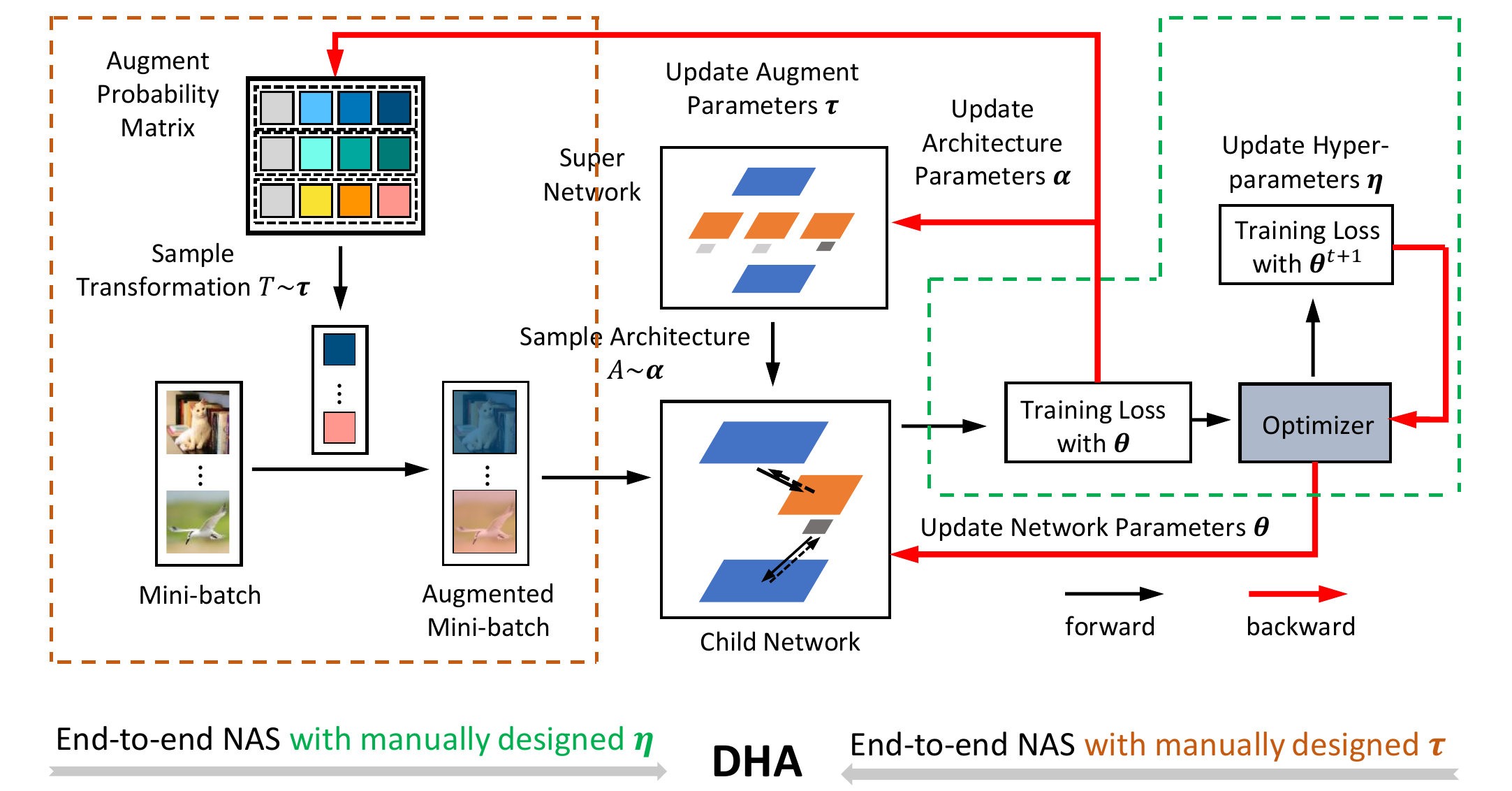}
    \caption{An overview of DHA. We first sample the DA operations for each sample based on the data transformation parameters $\bm{\tau}$. Then, a child network is sampled based on the architecture parameters $\bm{\alpha}$, which will be used to process the transformed mini-batch. Training loss is calculated to update the data transformation parameters $\bm{\tau}$, architecture parameters $\bm{\alpha}$, and neural network parameters $\bm{\theta}$. Then the training loss based on updated networks' weights $\bm{\theta}^{t+1}$ is used to update hyper-parameters $\bm{\eta}$.}
    \label{fig:overview}
\end{figure*}

\section{Related Works}
\noindent \textbf{Data augmentation.} Learning data augmentation policies for a target dataset automatically has become a trend, considering the difficulty to elaborately design augmentation methods for various datasets~\citep{cubuk2018autoaugment,zhang2019adversarial,lin2019online}.
Specifically, AutoAugment~\citep{cubuk2018autoaugment} and Adversarial AutoAugment~\citep{zhang2019adversarial} adopt reinforcement learning to train a controller to generate policies, while OHL-Auto-Aug~\citep{lin2019online} formulates augmentation policy as a probability distribution and adopts REINFORCE~\citep{williams1992simple} to optimize the distribution parameters along with network training. PBA~\citep{ho2019population} and FAA~\citep{lim2019fast} use population-based training method and Bayesian optimization respectively to reduce the computing cost of learning policies. \cite{cubuk2020randaugment} argues that the search space of policies used by these works can be reduced greatly and simple grid search can achieve competitive performance. They also point out that the optimal data augmentation policy depends on the model size, which indicates that fixing an augmentation policy when searching for neural architectures may lead to sub-optimal solutions.

\noindent \textbf{Hyper-parameter optimization.}
Hyper-parameters also play an important role in the training paradigm of deep neural network models.
Various black-box optimization approaches have been developed to address hyper-parameter tuning tasks involving multiple tasks \citep{mittal2020hyperstar, perrone2018scalable} or mixed variable types \citep{ru2019bayesian}. Meanwhile, techniques like multi-fidelity evaluations \citep{kandasamy2017multi,wu2019practical}, parallel computation \citep{Gonzalez_2015Batch, Kathuria_NIPS2016Batched, alvi2019asynchronous}, and transfer learning \citep{swersky2013multi, min2020generalizing} are also employed to further enhance the query efficiency of the hyper-parameter optimization. In addition, several works \citep{bengio2000gradient,lorraine2020optimizing, mackay2019self, shaban2019truncated, maclaurin2015gradient} have proposed to use gradient-based bi-level optimization to tune large number of hyper-parameters during model training. Although many HPO strategies have been adopted in NAS strategies, methods that jointly optimize both architectures and hyper-parameters are rarely seen except the ones discussed below. 

\noindent \textbf{Neural architecture search.} NAS has attracted growing attention over the recent years and provided architectures with better performance over those designed by human experts~\citep{pham2018efficient,Real2018_AmoebaNet,liu2018darts}. 
The rich collection of NAS literature can be divided into two categories: query-based methods and gradient-based ones. The former includes powerful optimization strategies such as reinforcement learning \citep{ZophLe17_NAS,pham2018efficient}, Bayesian optimization \citep{kandasamy2018neural, ru2020neural} and evolutionary algorithms \citep{elsken2018efficient,lu2019nsga}. 
The latter enables the use of gradients in updating both architecture parameters and network weights, which significantly reducing the computation costs of NAS via weight sharing~\citep{liu2018darts,chen2019progressive, xie2018snas, hu2020dsnas}. 
To reduce searching cost, most NAS methods search architectures in a low-fidelity set-up (e.g.fewer training epochs, smaller architectures) and retrain the optimal architecture using the full set-up before deployment. This separation of \textit{search} and \textit{evaluation} is sub-optimal~\citep{hu2020dsnas},
which motivates the development of end-to-end NAS strategies \citep{xie2018snas, hu2020dsnas} that return read-to-deploy networks at the end of the search. Our work also proposes an end-to-end solution.

\noindent \textbf{Joint optimization of AutoML components.}
Conventional neural architecture search methods perform a search over a fixed set of architecture candidates and then adopt a separate set of hyper-parameters when retraining the best architecture derived from the architecture search phase. Such search protocol may lead to sub-optimal results \citep{zela2018towards, dong2020autohas} as it neglects the influence of training hyper-parameters on architecture performance and ignores superior architectures under alternative hyper-parameter values \citep{dai2020fbnetv3}. Given this, several works have been proposed to jointly optimize architecture structure and training hyper-parameters \citep{dai2020fbnetv3, wang2020apq, dong2020autohas}. \cite{zela2018towards} introduces the use of multi-fidelity Bayesian optimization to search over both the architecture structure and training hyper-parameters. \cite{dai2020fbnetv3} trains an accuracy predictor to estimate the network performance based on both the architecture and training hyper-parameters and then uses an evolutionary algorithm to perform the search. Both these methods are 
query-based and require a relatively large number of architecture and hyper-parameter evaluations to fine-tune predictors or obtain good recommendations. 
To improve the joint optimization efficiency, AutoHAS~\citep{dong2020autohas} introduces a differentiable approach in conjunction with weight sharing for the joint optimization task, which empirically demonstrates that such a differentiable one-shot approach achieves superior efficiency over query-based methods. In addition to jointly optimizing neural architectures and hyper-parameters, the other line of research focuses on the joint optimization of neural architectures and data augmentation hyper-parameters~\citep{DBLP:journals/corr/abs-2012-09407}. 
Our proposed method differs from the above works in mainly two aspects: first, our method is more efficient than AutoHAS since our method has no need to update the whole network and only needs to update the sampled sub-network at each optimization step. Second, the joint optimization scope is further extended from NAS and training hyper-parameters to include DA, HPO and NAS.

\section{Methodology}
Consider a dataset $\mathcal{D} =\{(x_i,y_i)\}_{i=1}^{N}$, where $N$ is the size of this dataset, and $y_i$ is the label of the input sample $x_i$. We aim to train a neural network $f(\cdot)$, which can achieve the best accuracy on the test dataset $\mathcal{D}^{test}$. Multiple AutoML components are considered, including DA, HPO, and NAS. Let $\bm{\tau}$, $\bm{\eta}$, $\bm{\alpha}$, and $\bm{\theta}$ represent the data augmentation parameters, the hyper-parameters, the architecture parameters, and the objective neural network parameters, respectively. This problem can be formulated as
\begin{equation}
\begin{split}
   & argmin_{\bm{\tau}, \bm{\eta}, \bm{\alpha}, \bm{\theta}} \mathcal{L}(\bm{\tau}, \bm{\eta}, \bm{\alpha}, \bm{\theta}; \mathcal{D})\\
   s.t. &  \quad c_i(\bm{\alpha}) \leq C_i, i = 1, ..., \gamma,
\label{equ:whole_constrain}
\end{split}
\end{equation}
where $\mathcal{L}(\cdot)$ represents the loss function, $\mathcal{D}$ denotes the input data, $c_i(\cdot)$ refers to the resource cost (e.g., storage or computational cost) of the current architecture $\alpha$, which is restricted by the $i$-th resource constraints $C_i$, and $\gamma$ denotes the total number of resource constraints.
Considering the huge search space, it is challenging to achieve the joint optimization of $\bm{\tau}$, $\bm{\eta}$, $\bm{\alpha}$, and $\bm{\theta}$ within one-stage without parameter retraining. In this work, we propose to use the differentiable method to provide a computationally efficient solution. See Fig.~\ref{fig:overview} for an illustration.

\subsection{Data augmentation parameters}\label{sec:DA}
For every mini-batch of training data $\mathcal{B}^{tr}=\{(x_k,y_k)\}_{k=1}^{n^{tr}}$ with batch size $n^{tr}$, we conduct data augmentation to increase the diversity of the training data. We consider $K$ data augmentation operations, and each training sample is augmented by a transformation consisting of two successive operations~\citep{cubuk2018autoaugment, lim2019fast}. Each operation is associated with a magnitude that is uniformly sampled from $[0, 10]$.
The data augmentation parameter $\bm{\tau}$ represents a probability distribution over the augmentation transformations. 
For $t$-th iteration, we sample $n^{tr}$ transformations according to $\bm{\tau}^t$ with Gumbel-Softmax reparameterization~\citep{maddison2016concrete} and to generate the corresponding augmented samples in the batch.
Given a sampled architecture, the loss function for each augmented sample is denoted by $\mathcal{L}^{tr}(f(\bm{\alpha}^t, \bm{\theta}^t; \mathcal{T}_k(x_k)))$, where $\mathcal{T}_k$ represents the selected transformation. In order to relax $\bm{\tau}$ to be differentiable, we regard $p_k(\bm{\tau}^t)$, the probability of sampling the transformation $\mathcal{T}_k$, as an importance weight for the loss function of corresponding sample $\mathcal{L}^{tr}(f(\bm{\alpha}^t, \bm{\theta}^t; \mathcal{T}_k(x_k)))$. 
The objective of data augmentation is to minimize the following loss function:
\begin{equation}
\begin{split}
\mathcal{L}^{DA}(\bm{\tau}^t) = - \sum_{k=1}^{n^{tr}} p_k(\bm{\tau}^t) \mathcal{L}^{tr}(f(\bm{\alpha}^t, \bm{\theta}^t; \mathcal{T}_k(x_k))).
\label{equ:DA_loss}
\end{split}
\end{equation}
With this loss function, DHA intends to increase the sampling probability of those transformations that can generate samples with high \textbf{training loss}. By sampling such transformations, DHA can pay more attention to more aggressive DA strategies and increase model robustness against difficult samples~\citep{zhang2019adversarial}. However, blindly increasing the difficulty of samples may cause the \textbf{augment ambiguity} phenomenon~\citep{DBLP:journals/corr/abs-2003-11342}: augmented images may be far away from the majority of clean images, which could cause the under-fitting of model and deteriorate the learning process. Hence, besides optimizing the probability matrix of DA strategies, we randomly sample the magnitude of each chosen strategy from an \textbf{uniform distribution}, which can prevent learning heavy DA strategies: augmenting samples with large magnitude strategies. Moreover, instead of training a controller to generate adversarial augmentation policies via reinforcement learning~\citep{zhang2019adversarial} or training an extra teacher model to generate additional labels for augmented samples~\citep{DBLP:journals/corr/abs-2003-11342}, we search for the probability distribution of augmentation transformations directly via gradient-based optimization. In this way, the optimization of data augmentation is very efficient and hardly increases the computing cost.

\subsection{Hyper-parameters}
\label{sec:HPO}
As shown in Fig.~\ref{fig:overview},
given the batch of augmented training data $\{(\mathcal{T}_k(x_k),y_k)\}_{k=1}^{n^{tr}}$ and the sampled child network,
we need to optimize the differentiable hyper-parameters $\bm{\eta}$, such as learning rate and L2 regularization. At the training stage, we alternatively update $\bm{\theta}$ and $\bm{\eta}$. 
In $t$-th iteration, we can update $\bm{\theta}^{t}$ based on the gradient of the unweighted training loss $\mathcal{L}^{tr}(f(\bm{\alpha}^t, \bm{\theta}^t; \mathcal{B}^{tr})) = \frac{1}{n^{tr}}\sum_{k=1}^{n^{tr}}\mathcal{L}^{tr}(f(\bm{\alpha}^t, \bm{\theta}^t; \mathcal{T}_k(x_k)))$, which can be written as: 
\begin{equation}
    \bm{\theta}^{t+1} =  \textit{OP}(\bm{\theta}^{t}, \bm{\eta}^t, \nabla_{\bm{\theta}}\mathcal{L}^{tr}(f(\bm{\alpha}^t, \bm{\theta}^t; \mathcal{B}^{tr}))),
    \label{equ:HPO_loss1}
\end{equation}
where $\textit{OP}(\cdot)$ is the optimizer. To update the hyper-parameters $\bm{\eta}$, we regard $\bm{\theta}^{t+1}$ as a function of $\bm{\eta}$ and compute the \textbf{training loss} $\mathcal{L}^{tr}(f(\bm{\alpha}^t, \bm{\theta}^{t+1}(\bm{\eta}^t); \mathcal{B}^{tr}))$ with network parameters $\bm{\theta}^{t+1}(\bm{\eta}^t)$ on a mini-batch of training data $\mathcal{B}^{tr}$. Then, $\bm{\eta}^t$ is updated with $\nabla_{\bm{\eta}}\mathcal{L}^{tr}(f(\bm{\alpha}^t, \bm{\theta}^{t+1}(\bm{\eta}^t); \mathcal{B}^{tr}))$ by gradient descent: 
\begin{equation}
\bm{\eta}^{t+1} = \bm{\eta}^t - \beta \nabla_{\bm{\eta}}\mathcal{L}^{tr}(f(\bm{\alpha}^t, \bm{\theta}^{t+1}(\bm{\eta}^t); \mathcal{B}^{tr})),
\label{equ:HPO_loss2}
\end{equation}
where $\beta$ is a learning rate. Even $\bm{\theta}^t$ can also be deployed to $\bm{\theta}^{t-1}$ whose calculation also involves $\bm{\eta}^{t-1}$, we take an approximation method in Eqn.~(\ref{equ:HPO_loss2}) and regard $\bm{\theta}^t$ here as a variable independent of $\bm{\eta}^{t-1}$. Instead of splitting an extra validation set for HPO, we directly sample a subset from training set to update $\bm{\eta}$, which could ensure that the whole training set is used in updating $\bm{\tau}$, $\bm{\eta}$, $\bm{\alpha}$, and $\bm{\theta}$.
This avoids the final learned weight decay coefficient to be zero as non-zero weight decay coefficient can help avoid the model to overfit to the training data.

\subsection{Architecture parameters}\label{NAS}
With the augmented data in previous section, we achieve the optimization of the architecture parameter $\bm{\alpha}$  through end-to-end NAS, motivated by SNAS \citep{xie2018snas}, DSNAS \citep{hu2020dsnas} and ISTA-NAS~\citep{yang2020ista}. Following \cite{liu2018darts}, we denote the each space as a single directed acyclic graph (DAG), where the probability matrix $\bm{\alpha}$ consists of vector $\bm{\alpha}_{i,j}^T=[\alpha_{i,j}^1,...,\alpha_{i,j}^r,...,\alpha_{i,j}^k]$ and $\alpha_{i,j}^r$ represents the probability of choosing $r^{th}$ operation associated with the edge $(i,j)$. Instead of directly optimizing $\bm{\alpha}\in\mathbb{R}^n$, we adopt ISTA-NAS to optimize its compressed representation $\mathbf{b}\in\mathbb{R}^m$ where $m<<n$, which can be written as:
\begin{equation}
    \mathbf{b} = \mathbf{A} \bm{\alpha} + \epsilon,
\label{architecture_loss}
\end{equation}
where $\epsilon\in\mathbb{R}^m$ represents the noise and $\mathbf{A} \in \mathbb{R}^{m\times n}$ is the measurement matrix which is randomly initialized. Eqn.~(\ref{architecture_loss}) is solved through using LASSO loss function \citep{tibshirani1996regression} and the $\bm{\alpha}$ is optimized by using iterative shrinkage thresholding algorithm \citep{daubechies2004iterative}, which can be written as:
\begin{equation}
\bm{\alpha}^{t+1} = \eta_{\lambda/L}(\bm{\alpha}^{t} - 
\frac{1}{L}\mathbf{A}^T(\mathbf{A} \bm{\alpha}-\mathbf{b})), t=0,1,...,
\label{architecture_loss2}
\end{equation}
where $L$ represents the LASSO formulation which can be written as $\mathop{min}\limits_{\bm{\alpha}} \frac{1}{2} ||\mathbf{A}\bm{\alpha}-\mathbf{b}||^2_2 + \lambda ||\bm{\alpha}||_1 $; the $\lambda$ represents the regularization parameters and the $\eta_{\lambda/L}$ is the shrinkage operator as defined in \citep{beck2009fast}. Thus we have:
\begin{equation}
        \bm{\alpha}_{j}^T \bm{o}_{j} = (\mathbf{b}_j^T\mathbf{A}_j-[\bm{\alpha}_j(\mathbf{b}_j)]^T\mathbf{E}_j)\mathbf{o}_j,
\label{achitecture_loss3}
\end{equation}
where $\mathbf{o}_{j}$ refers to all possible operations connected to note $j$ and $\mathbf{E}_j =\mathbf{A}_j^T\mathbf{A}_j - \mathbf{I} $. With this relaxation, $\mathbf{b}$ can be optimized through calculating the gradient concerning \textbf{training loss}. The main reason for using this optimization algorithm is that it can optimize the high-dimensional architecture parameters through optimizing low-dimensional embeddings, which can largely decrease the optimization difficulty and increase the optimization efficiency. Moreover, this algorithm also adopt the weights sharing in the optimization process which can be readily combined with our proposed data augmentation and hyper-parameter optimization method.

\subsection{Joint-optimization}

\begin{algorithm}[t] 
\caption{DHA} 
\label{algo:whole_algo}
\hspace*{0.02in} {\bf Initialization:}
Data Transformation Parameters $\bm{\tau}$, Hyper-parameters $\bm{\eta}$, Compressed Representation $\mathbf{b}$, Measurement Matrix $\mathbf{A}$, and Network Parameters $\bm{\theta}$ \\
\hspace*{0.02in} {\bf Input:}
Training Set $\mathcal{D}^{tr}$, Parameters $\bm{\tau}$, $\bm{\eta}$, $\mathbf{b}$, $\mathbf{A}$, $\bm{\theta}$, and the iteration number $T$\\
\hspace*{0.02in} {\bf Return:}
$\bm{\tau}, \bm{\eta}, \bm{\alpha}, \bm{\theta}$
\begin{algorithmic}[1] 
\While{$t<T$}
    \State Separately sample a mini-batch $\mathcal{B}^{tr}$ from $\mathcal{D}^{tr}$;
    \State For each sample $x_k$ in mini-batch $\mathcal{B}^{tr}$, sample a transformation $\mathcal{T}_k(x_k)$ according to $\bm{\tau}^t$;
    \State Recover $\bm{\alpha}^t$ by solving Eqn.~(\ref{architecture_loss}) with $\mathbf{b}^t$ and $\mathbf{A}$; 
    \State Extract a child network from the super network;
    \State Compute the weighted training loss function as Eqn.~(\ref{equ:DA_loss}) and update $\bm{\tau}^{t+1}$ accordingly; \label{update_T}
    \State Calculate $\bm{\theta}^{t+1}$ with Eqn.~(\ref{equ:HPO_loss1});
    \State Use training loss function to update $\mathbf{b}^{t+1}$ through the gradient descent, then $\bm{\alpha}^{t+1}$ is updated with Eqn.~(\ref{architecture_loss2});\label{update_A}
    \State Compute the training loss function with $\bm{\theta}^{t+1}$ on $\mathcal{D}^{val}$ and update $\bm{\eta}^{t+1}$ with Eqn.~(\ref{equ:HPO_loss2});\label{update_H}
\EndWhile
\end{algorithmic}
\end{algorithm}
\vspace{-0.4cm}

Based on the above analysis of each AutoML module, DHA realizes end-to-end joint optimization of automated data augmentation parameters $\bm{\tau}$, hyper-parameters $\bm{\eta}$, and architecture parameters $\bm{\alpha}$. The DHA algorithm is summarized in Algorithm~\ref{algo:whole_algo}. One-level optimization is applied to $\bm{\tau}$ and $\bm{\alpha}$ as in Line 6 and Line 8, while bi-level optimization is applied to $\bm{\eta}$ as in Line 9. 

One thing worth mentioning is that different optimizers are adopted for different parameters. There are two main reasons behind this choice. Firstly, $\bm{\tau}, \bm{\eta}, \bm{\alpha}$ and $\bm{\theta}$ work differently in the optimization process, e.g., $\bm{\tau}$ controls the transformation strategy for the training set while $\bm{\alpha}$ is related to the architecture selection. Besides $\bm{\theta}$, other parameters could not directly be optimized through the gradient descent on training set. These differences cause the different optimization methods for $\bm{\tau}, \bm{\alpha}, \bm{\eta}$ and $\bm{\theta}$. Secondly, $\bm{\tau}, \bm{\alpha}, \bm{\eta}$ and $\bm{\theta}$ have different dimensions and different scales, which makes the joint-optimization with a uniform optimizer extremely impracticable. This is why current works concerning DA~\citep{ho2019population}, NAS~\citep{yao2020sm,nekrasov2019fast} and HPO~\citep{falkner2018bohb} always adopt different optimizer for network's weights and hyper-parameters they aim to optimize. During the training process, DA and HPO adapt the online optimization strategy. The DA strategy and HPO settings are evolving with the weights of the parameters.  

Moreover, the main reason that DHA could optimize large-scale search space in an effectively manner, is that DHA delicately adopt  \textbf{weight-sharing} in the joint-optimization for different parameters. Instead of only optimizing a sub-network with a DA strategy and a hyper-parameter setting to check the performance of certain setting, we realize the joint-optimization with the help of a super-net network, a DA probability matrix and continuous hyper-parameter setting. In that way, DHA can make use of previous trained parameter weights to check the performance setting, which largely decreases the computational request.

\begin{figure*}[t]
\setlength{\abovecaptionskip}{0.1cm}
\setlength{\belowcaptionskip}{-0.5cm}
    \centering
    \includegraphics[width=13.0cm]{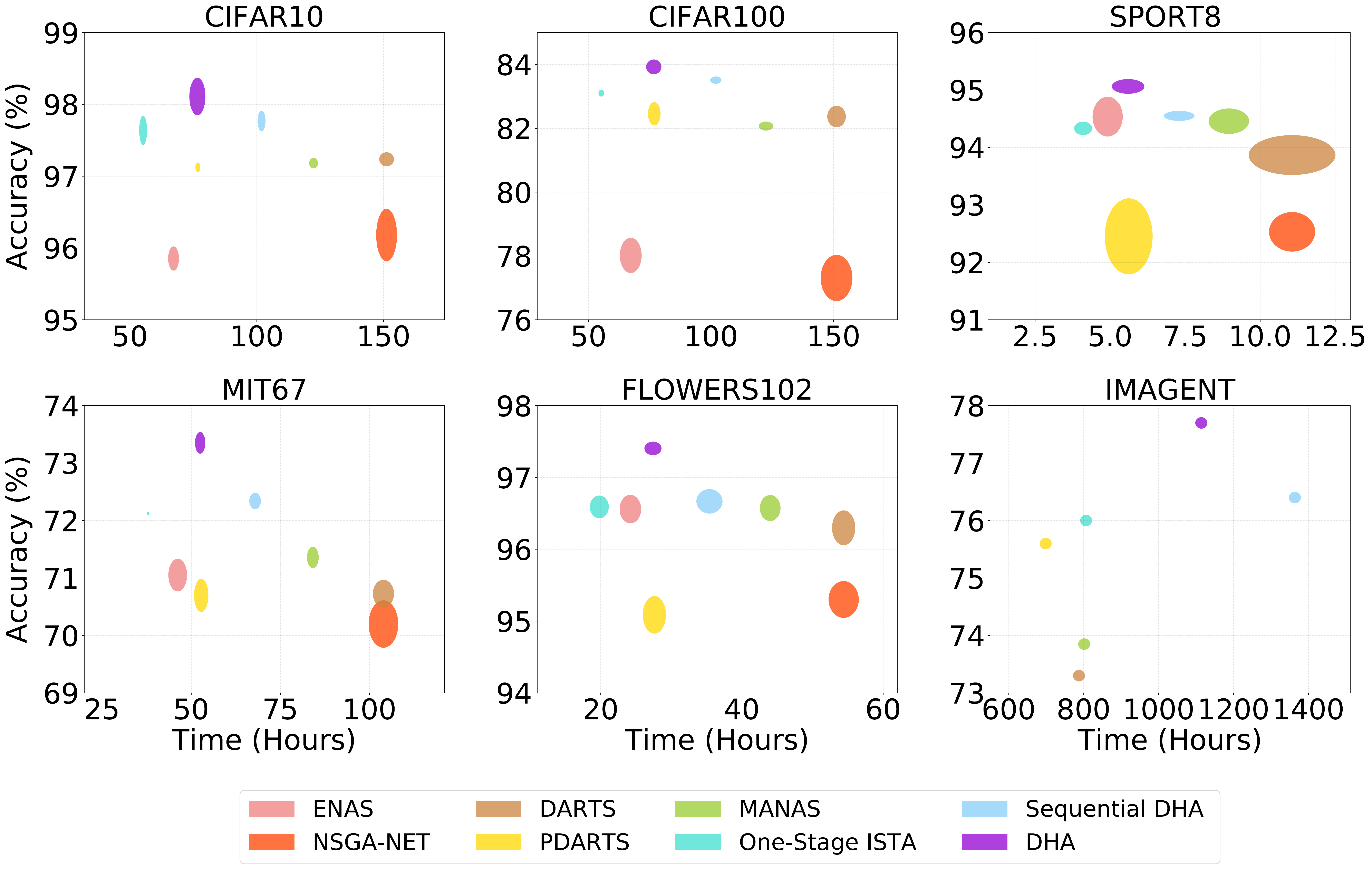}
    \caption{Top-1 accuracy and computational time of AutoML algorithms for classification task on CIFAR10, CIFAR100, SPORT8, MIT67, FLOWERS102 and ImageNet. DHA is conducted with Cell-Based search space. Ellipse centres and ellipse edges represent the $\mu \pm \{ 0, \sigma/2 \}$, respectively (mean $\mu$, standard deviation $\sigma$). For ImageNet, we present test accuracy without error bars, as the error bars are not reported in existing works.}
    \label{fig:performance comparison}
\end{figure*}

\begin{table*}[hbt!]
\caption{Top-1 accuracy ($\%$) and computational time (GPU hour) of different AutoML algorithms on CIFAR10, CIFAR100, SPORT8, MIT67 and FLOWERS102 with Cell-Based Search Space.
We report the sum of search time and tune time for two-stage NAS and the whole running time for one-stage NAS.
All the methods are implemented by ourselves.
Different NAS algorithms for one dataset are performed under the similar parameter weights constrain to ensure fair comparison.}
\setlength{\belowcaptionskip}{-0.2cm}
\begin{center}
\scalebox{0.92}{\begin{tabular}{c|c|c|c|c|c}
\hline
\hline
\multirow{2}*{\textbf{Model}} & \textbf{CIFAR10} & \textbf{CIFAR100} & \textbf{SPORT8} & \textbf{MIT67} & \textbf{FLOWERS102} \\
& \textbf{Acc} & \textbf{Acc} &   \textbf{Acc} & \textbf{Acc}  & \textbf{Acc} \\
\hline
\hline
ENAS~\citep{pham2018efficient}           & 95.85±0.17& 78.02±0.55 & 94.54±0.35  & 71.05±0.29& 96.56±0.20 \\
NSGA-NET~\citep{lu2019nsga}       & 96.18±0.37 & 77.31±0.14 & 92.53±0.34 & 70.20±0.41 & 95.30±0.26 \\
DARTS~\citep{liu2018darts}          & 97.24±0.10 & 82.37±0.34 & 93.87±0.35 & 70.73±0.24& 96.30±0.24 \\
P-DARTS~\citep{chen2019progressive}         & 97.13±0.07& 82.46±0.37 & 92.45±0.66  & 70.70±0.29 & 95.09±0.26\\
MANAS~\citep{carlucci2019manas}          & 97.18±0.07& 82.07±0.14& 94.46±0.22  & 71.36±0.19  & 96.57±0.18 \\
One-Stage ISTA~\citep{yang2020ista} & 97.64±0.20 & 83.10±0.11& 94.33±0.12 & 72.12±0.03 & 96.59±0.16\\
Sequential DHA & 97.77±0.14 & 83.51±0.12 & 94.55±0.09& 72.34±0.14  & 96.67±0.17 \\
DHA  & \textbf{98.11}±0.26  & \textbf{83.93}±0.23 & \textbf{95.06}±0.13 & \textbf{73.35}±0.19 & \textbf{97.41}±0.09\\
\hline
\hline
& \textbf{Time} & \textbf{Time} &   \textbf{Time} & \textbf{Time}  & \textbf{Time} \\
\hline
\hline
ENAS~\citep{pham2018efficient} & 67.2±2.1 & 67.2±4.4 & 4.92±0.5 & 46.2±2.6 & 24.2±1.5 \\
NSGA-NET~\citep{lu2019nsga} &151.2±4.1 & 151.2±6.5 & 11.06±0.8 & 103.9±4.1 & 54.4±2.1 \\
DARTS~\citep{liu2018darts} & 151.2±2.9 & 151.2±3.8 & 11.06±1.4 & 103.9±3 & 54.4±1.6 \\
P-DARTS~\citep{chen2019progressive}  & 76.8±1 & 76.8±2.6 & 5.62±0.8 & 52.8±2 & 27.6±1.6 \\
MANAS~\citep{carlucci2019manas} & 122.4±1.8 & 122.4±2.9 & 8.96±0.7 & 84.1±1.6 & 44±1.5 \\
One-Stage ISTA~\citep{yang2020ista} & 55.2±1.5 & 55.2±1.2 & 4.1±0.3 & 37.9±0.4 & 19.8±1.3 \\
Sequential DHA  & 101.9±1.6 & 101.9±2.3 & 7.3±0.5 & 67.9±1.6 & 35.4±1.9 \\
DHA & 76.6±3.2 & 76.6±3.1 & 5.6±0.5 & 52.5±1.4 & 27.4±1.2 \\
\hline
\hline
\end{tabular}}
\label{table:MAIN_RESULT}
\end{center}
\end{table*} 

\section{Experiments}
\label{sec:exp}
In this section, we empirically compare DHA against existing AutoML algorithms on various datasets.
With extensive experiments, we demonstrate the benefits of joint-optimization over sequential-optimization in terms of generalization performance and computational efficiency. 

\subsection{Experiment setting}
\label{sec: exp setting}

\textbf{Datasets.} Following~\cite{ru2020neural}, we conducted experiments on various datasets, including CIFAR10 and CIFAR100 for the object classification task \citep{krizhevsky2009learning}, SPORT8 for the action classification task \citep{li2007and}, MIT67 for the scene classification task \citep{quattoni2009recognizing}, FLOWERS102 for the fine object classification task \citep{nilsback2008automated}, and ImageNet\citep{russakovsky2015imagenet} for the large-scale classification task. The accuracy is calculated on the test set. 

\noindent \textbf{Search space.} \textbf{(1) Automated DA.} Following~\cite{ho2019population},
we consider $K = 14$ different operations for data augmentation, such as AutoContrast and Equalize. The magnitude of each operation is randomly sampled from the uniform distribution.
Note that we follow a widely used data augmentation setting~\citep{zhang2019adversarial, zhou2021metaaugment}, which applies two successive data augmentation strategies on each training sample. Our method can be easily extended to more general cases by modifying the probability matrix and the range of sampling distribution. For example, for the case of three successive data augmentation strategies, we could extend the size of probability matrix from $K^2$ to $K^3$, where $K$ is the number of data augmentation categories.
\textbf{(2) NAS}. Following \cite{liu2018darts} and \cite{cai2018proxylessnas}, we consider both the cell-based and the MobileNet search space, which regards the whole architecture as a stack similar cells. 
\textbf{(3) HPO.} We consider both the L2 regularization (i.e., weight decay) and the learning rate in the experiments.
Detailed information is provided in Appendix~\ref{app:experimental setting}.

\noindent \textbf{Baselines.} We compare DHA with various AutoML algorithms, such as ENAS~\citep{pham2018efficient}, NSGA-NET~\citep{lu2019nsga} , NSGA-NET~\citep{lu2019nsga} , P-DARTS~\citep{chen2019progressive}, MANAS~\citep{carlucci2019manas}  and One-Stage ISTA~\citep{yang2020ista}(see Table~\ref{table:MAIN_RESULT} and Table~\ref{table:MAIN_RESULT R1}).
For datasets including CIFAR10, CIFAR100, SPORT8, MIT67, and FLOWERS102, we re-implement the baseline methods. Specifically, for papers with code, incluing ENAS, NSGA-NET, P-DARTS and One-Stage ISTA, we directly use their official implementation code. For paper without code, including DARTS and MANAS, we use the unofficial re-implementation code which achieves similar performance described in the original papers.
As for ImageNet, we directly refer to the model performance reported in papers cited in Table~\ref{table:MAIN_RESULT R1}.
To further demonstrate the benefits of joint optimization of multiple AutoML components, we also include a baseline, \textbf{Sequential DHA}, which resembles the common practice by human to optimize different components in sequence. Specifically, Sequential DHA consists of two stages. During the first stage, Sequential DHA performs NAS to find the optimal architecture under certain hyper-parameter settings. In the next stage, Sequential DHA performs the online DA and HPO strategy proposed in our paper and trains the architecture derived from the first stage from scratch. 
Detailed algorithm of Sequential DHA could be found in Appendix~\ref{app:sequential DHA}.

\noindent \textbf{Implementation details.}
For hyper-parameters of DHA, we simply adopt the hyper-parameters used in previous works without many modifications, including Meta-Aug~\citep{zhou2021metaaugment}, ISTA-NAS~\citep{yang2020ista}, and DSNAS~\citep{hu2020dsnas}.
Experiments are conducted on NVIDIA V100 under PyTorch-1.3.0 and Python 3.6. Detailed settings of baselines are provided in Appendix~\ref{app:experimental setting}.

\subsection{Results}
\label{search_experiment}

\begin{table*}[hbt!]
\centering
\caption{Comparison with SOTA image classifiers on ImageNet in the Cell-Based (C) setting or MobileNet/ShuffleNet (M) setting. ($^{\dagger}$ denotes the architecture is searched on ImageNet, otherwise it is searched on CIFAR-10 or CIFAR-100.)}
\scalebox{0.82}{
\begin{tabular}{c|cc|c|cc|c|c}
\hline\hline
\multirow{2}*{\textbf{Model}} & \multicolumn{2}{c|}{\textbf{Test Acc (\%)}}  & \textbf{Params} & \multicolumn{2}{c|}{\textbf{Cost (GPU-day)}} & \multirow{2}*{\textbf{Search Attribute}} & \textbf{Search}\\ 
\cline{2-3}\cline{5-6}
 & Top-1 & Top-5 & \textbf{(M)} & Search & Eval & & \textbf{Space} \\ 
\hline\hline
DARTS (2nd)~\citep{liu2018darts} & 73.3 & 91.3 & 4.7 & 4.0 & 3.6 $\times$ 8  & Arch  & C \\
SNAS (mild)~\citep{xie2018snas} & 72.7 & 90.8 & 4.3 & 1.5 & 3.3 $\times$ 8 & Arch & C \\
GDAS~\citep{dong2019search} & 74.0 & 91.5 & 5.3 & 0.3 & 3.6 $\times$ 8 & Arch & C \\
BayesNAS~\citep{zhou2019bayesnas} & 73.5 & 91.1 & 3.9 & 0.2 & 3.6 $\times$ 8 & Arch  & C \\
PARSEC~\citep{casale2019probabilistic} & 74.0 & 91.6 & 5.6 & 1.0 & 3.6 $\times$ 8 & Arch & C \\
P-DARTS (CIFAR-10)~\citep{chen2019progressive} & 75.6 & 92.6 & 4.9 & 0.3 & 3.6 $\times$ 8 & Arch & C\\
P-DARTS (CIFAR-100)~\citep{chen2019progressive} & 75.3 & 92.5 & 5.1 & 0.3 & 3.6 $\times$ 8 & Arch & C \\
PC-DARTS (ImageNet)~\citep{xu2019pc}$^{\dagger}$ & 75.8 & 92.7 & 5.3 & 3.8 & 3.9 $\times$ 8 & Arch & C\\
GAEA+PC-DARTS~\citep{li2020geometry}$^{\dagger}$ & 76.0 & 92.7 & 5.6 & 3.8 & 3.9 $\times$ 8 & Arch & C\\
DrNAS~\citep{chen2020drnas}$^{\dagger}$ & 76.3 & 92.9 & 5.7 & 4.6 &  3.9 $\times$ 8 & Arch & C\\
Two-Stage ISTA~\citep{yang2020ista}$^{\dagger}$ & 75.0 & 91.9 & 5.3 & 2.3 & 3.4 $\times$ 8 & Arch & C \\
One-Stage ISTA~\citep{yang2020ista}$^{\dagger}$ & 76.0 & 92.9 & 5.7 & \multicolumn{2}{c|}{4.2 $\times$ 8} & Arch & C \\
\cline{1-8}
EfficientNet-B0~\citep{tan2019efficientnet}$^{\dagger}$ & 77.1 & 93.3 & 5.3 & - & - & Arch & M\\
SinglePathNAS~\citep{guo2019single}$^{\dagger}$ & 74.7 & - & 3.4 & 13.0 & 2.0 $\times$ 8 & Arch & M\\
ProxylessNAS (GPU)~\citep{cai2018proxylessnas}$^{\dagger}$ & 75.1 & 92.5 & 7.1 & 8.3 & 3.6 $\times$ 8 & Arch & M\\
DSNAS~\citep{hu2020dsnas}$^{\dagger}$ & 74.3 & 91.9 & - &  \multicolumn{2}{c|}{3.7 $\times$ 8} & Arch & M\\
OFA (small)~\citep{cai2019once}$^{\dagger}$ & 76.9 & 93.3 & 5.8 & \multicolumn{2}{c|}{6.8 $\times$ 8} & Arch + Resolution & M\\ 
APQ~\citep{wang2020apq}$^{\dagger}$ & 75.1 & - & - & \multicolumn{2}{c|}{12.5 $\times$ 8} & Arch + Pruning & M \\ 
AutoHAS \citep{dong2020autohas} $^{\dagger}$ & 74.2 & - & - & - &  - & Arch + HPO & M \\
\hline\hline
Sequential DHA$^{\dagger}$ & 76.7 & 93.8 & 5.4 & 3.2 &  5.5 $\times$ 8 &
Arch + DA + HPO & C  \\
DHA$^{\dagger}$ & 77.4 & 94.6 & 5.6 & \multicolumn{2}{c|}{5.8 $\times$ 8}  & Arch + DA + HPO & C  \\
\cline{1-8}
Sequential DHA$^{\dagger}$ & 77.1 & 93.9 & 5.7 & 5.8 &  5.7 $\times$ 8 &
Arch + DA + HPO & M  \\
DHA$^{\dagger}$ & \textbf{77.6} & 94.8 & 5.3 & \multicolumn{2}{c|}{5.4 $\times$ 8}  & Arch + DA + HPO & M \\
\hline\hline
\end{tabular}}
\label{table:MAIN_RESULT R1}
\end{table*} 

\begin{table*}[t]
\caption{Top-1 accuracy ($\%$) and computational time (GPU hour) of different combination of AutoML components on CIFAR10, CIFAR100, SPORT8, MIT67 and FLOWERS102. Listed algorithms are described in Ablation study.}
\begin{center}
\scalebox{0.82}{\begin{tabular}{c|cc|cc|cc|cc|cc}
\hline
\hline
\multirow{2}*{\textbf{Model}} & \multicolumn{2}{c|}{\textbf{CIFAR10}} & \multicolumn{2}{c|}{\textbf{CIFAR100}} & \multicolumn{2}{c|}{\textbf{SPORT8}} & \multicolumn{2}{c|}{\textbf{MIT67}} & \multicolumn{2}{c}{\textbf{FLOWERS102}} \\
\cline{2-11}
& \textbf{Acc} & \textbf{Time}  & \textbf{Acc} & \textbf{Time} &  \textbf{Acc} & \textbf{Time} & \textbf{Acc} & \textbf{Time}  & \textbf{Acc} & \textbf{Time} \\ 
\hline
\hline
Sequential NAS+DA          & 97.74 & 57.50  & 83.45 & 57.50  & 94.49 & 4.20 & 72.22 & 39.20 & 96.64 & 20.50 \\
Sequential NAS+HPO         & 97.54 & 63.60  & 83.13 & 63.60  & 94.47 & 4.60 & 72.10  & 43.10 & 96.57 & 22.50 \\
Sequential DHA             & 97.77 & 101.90 & 83.51 & 101.90 & 94.55 & 7.30 & 72.34 & 67.90 & 96.67 & 35.40 \\
\hline
Joint-optimization NAS+DA  & 97.78 & 64.00  & 83.55 & 64.00  & 94.53 & 4.70 & 72.38 & 43.80 & 96.72 & 22.90 \\
Joint-optimization NAS+HPO & 97.75  & 70.80  & 83.23 & 70.80  & 94.50 & 5.20 & 72.23 & 47.80 & 96.69 & 25.40 \\
DHA                        & \textbf{98.11} & 76.60  & \textbf{83.93} & 76.60  & \textbf{95.06} & 5.60 & \textbf{73.35} & 52.50 & \textbf{97.41} & 27.40 \\
\hline
\hline
\end{tabular}} 
\label{table:ABLATION1}
\end{center}
\end{table*} 

The test accuracy and computational time of various AutoML algorithms are summarized in Table~\ref{table:MAIN_RESULT} and Table~\ref{table:MAIN_RESULT R1}.
The timing results in these two tables measure the computational time taken to obtain a ready-to-deploy network, which corresponds to the sum of search and retrain time for two-stage NAS methods and the search time for end-to-end methods like one-stage NAS method as well as our DHA. 

\noindent \textbf{Small-scale datasets.}  As shown in Table~\ref{table:MAIN_RESULT}, methods optimizing all of DA, HPO and NAS automatically (i.e, Sequential DHA and DHA) consistently outperform those NAS algorithms with manual designed DA and HPO. Specifically, DHA achieves SOTA results on all datasets. This shows the clear performance gain of extending the search scope from architecture to including also data augmentation and hyper-parameters, justifying the need for multi-component optimization in AutoML. 
Moreover, despite optimising over a larger search space, DHA remains cost efficiency. For example, on CIFAR100, DHA enjoys $1.56\%$ higher test accuracy than DARTS but requires $42\%$ less time. Besides, the comparison between DHA and Sequential DHA reveals the evident advantage of doing DA, HPO and NAS jointly over doing them separately in different stages.

\begin{figure}[h]
\setlength{\abovecaptionskip}{0.5cm}
\setlength{\belowcaptionskip}{-0.15cm}
    \centering
    \includegraphics[width=9.0cm]{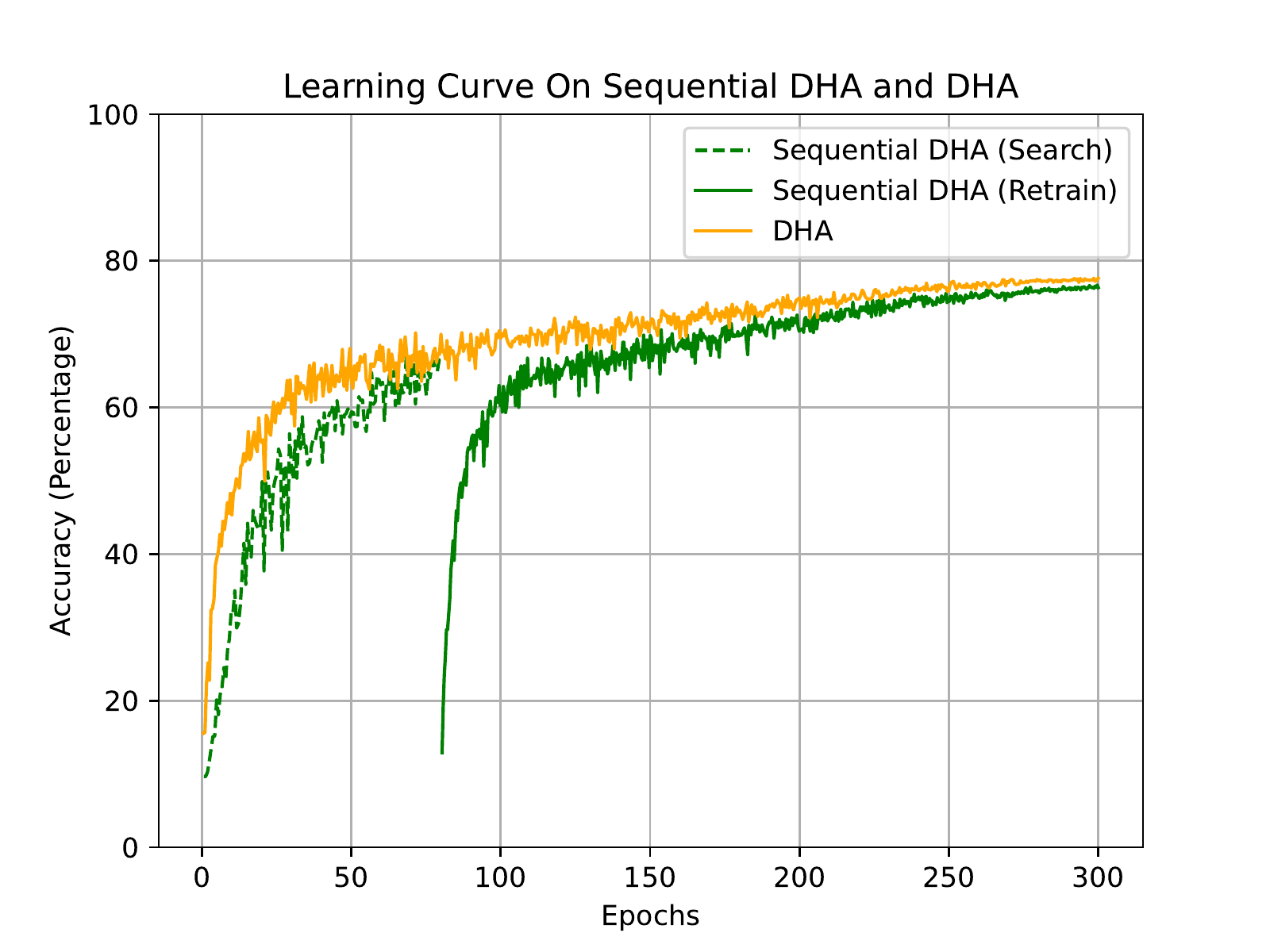}
    \caption{The validation accuracy of Sequential-DHA and DHA over the training time on ImageNet with Cell-Based Structure.}
    \label{fig:learning_curve}
\end{figure}

\noindent \textbf{Large-scale dataset.} Results of the large-scale dataset ImageNet with cell-based search space and MobileNet search space are shown in Table~\ref{table:MAIN_RESULT R1}. DHA consistently outperforms various NAS methods which only involves architecture optimization, demonstrating the benefits of joint-optimization. Even when compared with One-stage NAS methods like ISTA, DHA achieves up to $1.7\%$ TOP-1 accuracy improvement. Moreover, in comparison with the joint-optimization algorithm APQ~\citep{wang2020apq} and AutoHAS~\citep{dong2020autohas}, DHA outperforms APQ by $2.5\%$ and outperforms AutoHAS by $4.3\%$. 
These comparisons reveal that DHA proposes an efficient and high-performed joint-optimization algorithm.
The Top-1 accuracy and computation time of these AutoML algorithms are also summarized in Fig.~\ref{fig:performance comparison}.
As can be seen, DHA consistently gains highest test accuracy on all five datasets while being more cost efficient than NAS methods and Sequential DHA.
The validation accuracy of Sequential-DHA and DHA during the training is shown in Fig.~\ref{fig:learning_curve}. We could notice that a traditional sequential optimization algorithm would disrupt the learning process, as it has to retrain the model after the searching phase. While benefiting from the end-to-end optimization process, DHA would have a smoother learning curve and also achieve better performance.

\subsection{Analysis of loss landscape}
\vspace{-0.01cm}

\begin{figure}[hbt!]
\setlength{\abovecaptionskip}{-0.1cm}
\setlength{\belowcaptionskip}{-0.5cm}
    \centering
    \subfigure[One-Stage ISTA]{\includegraphics[width = 4.8cm]{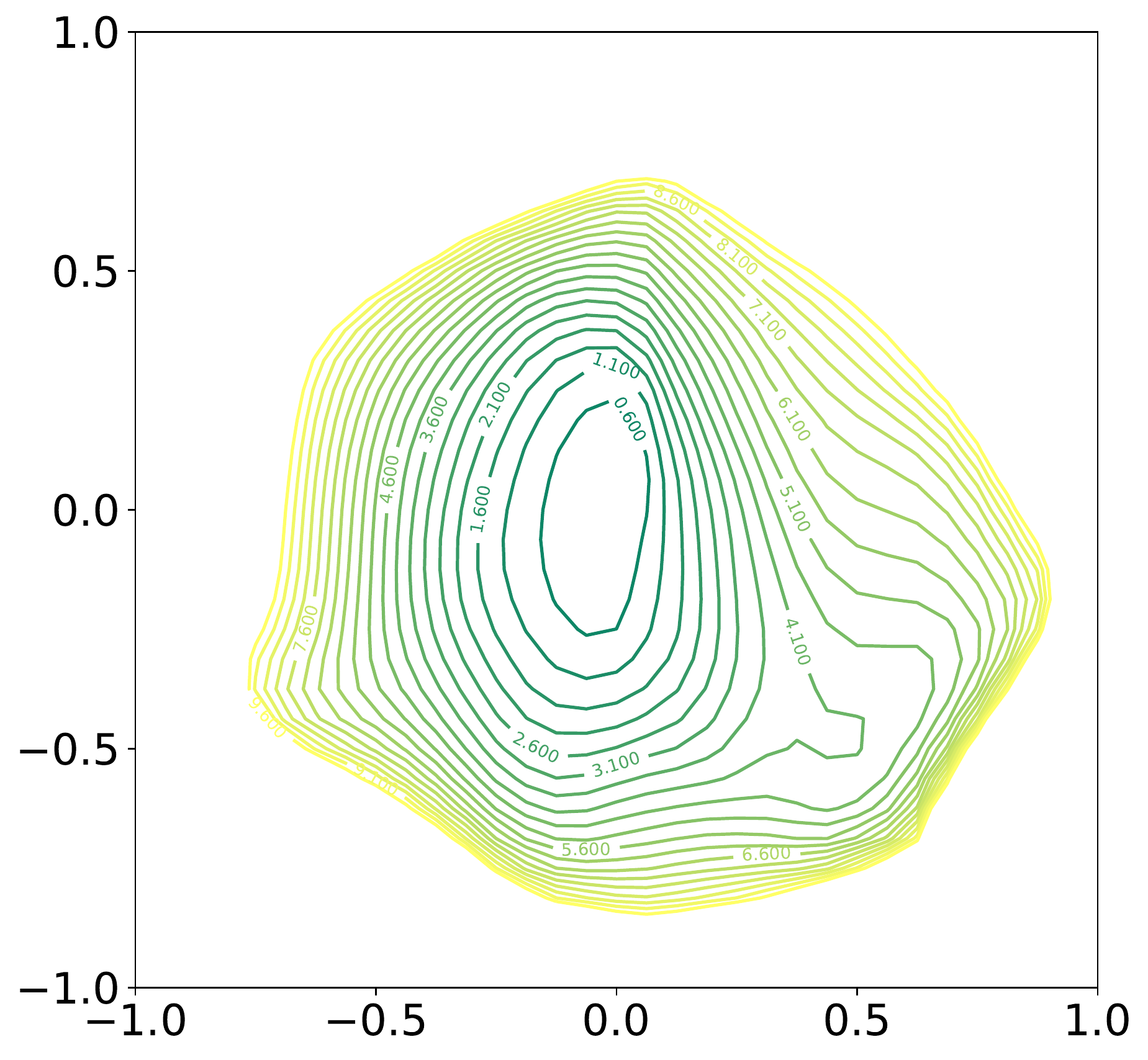}}
    \subfigure[Sequential DHA]{\includegraphics[width = 4.8cm]{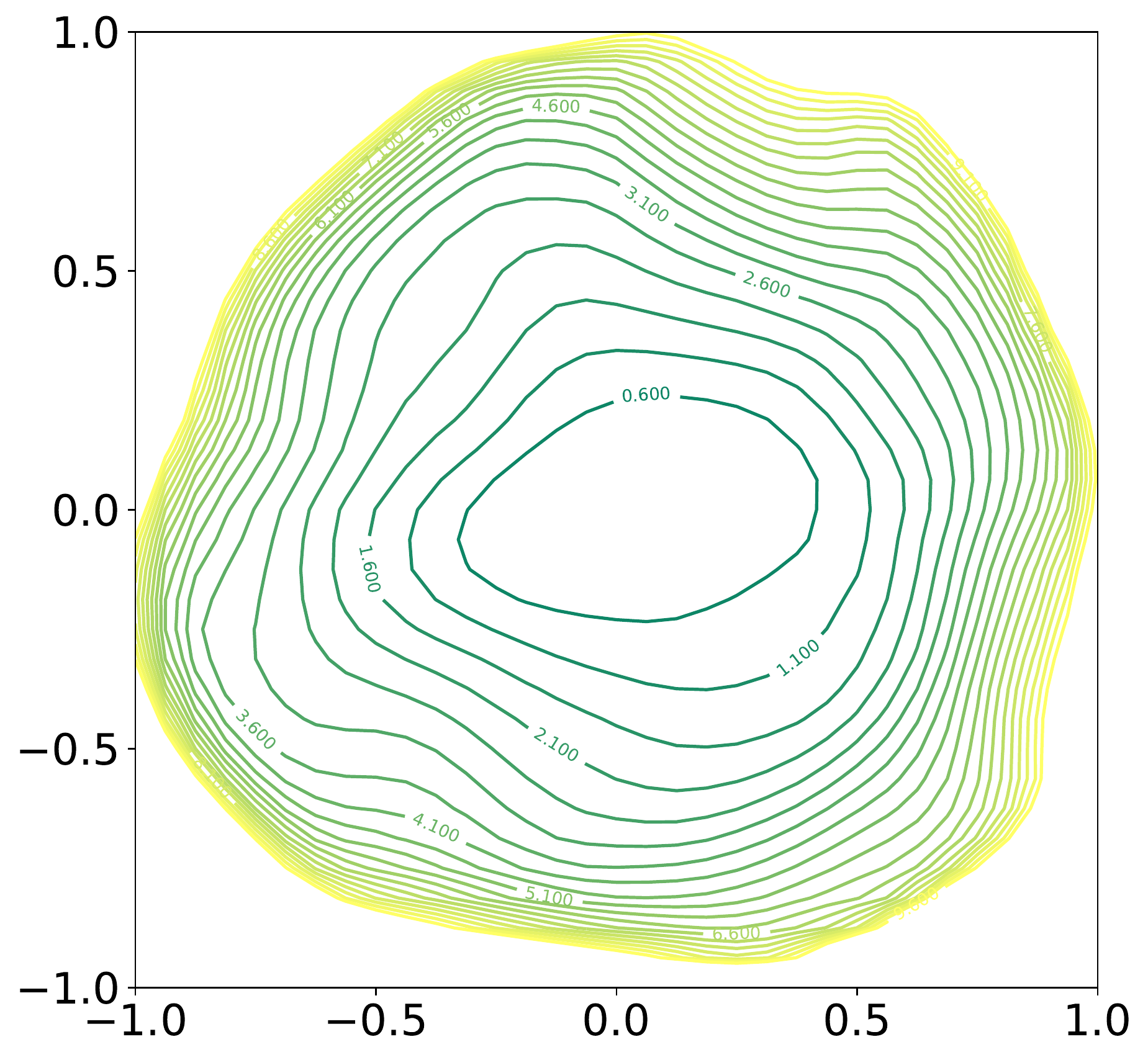}}
    \subfigure[DHA]{\includegraphics[width = 4.8cm]{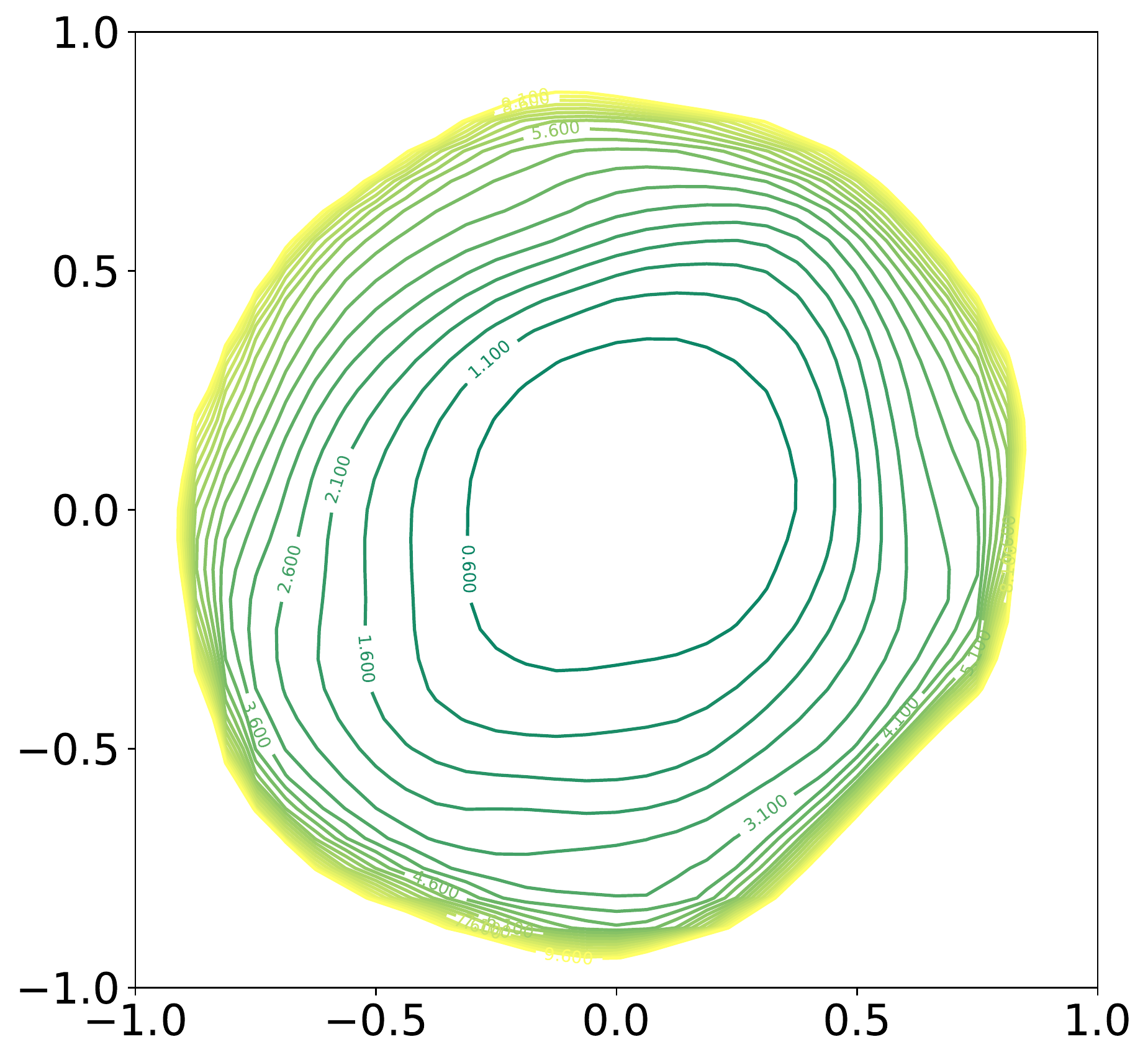}}
    \caption{Loss-landscape of the trained models on SPORT8. (a) One-Stage ISTA (b) Sequential DHA and (c) DHA.}
    \label{fig:LOSS_LANDSCAPE}
\end{figure}

The strong test performance of DHA across various datasets motivates us to further check the geometry of the minimiser achieved by our final trained model. 
We employ the filter-normalisation method proposed in \cite{li2017visualizing} to visualise the local loss landscape around the minimiser achieved by One-Stage ISTA, Sequential DHA and DHA. We choose a center point $\bm{\theta}^*$ in the graph, and two direction vector $\bm{\delta}$ and $\bm{\eta}$. We the plot the 2D contour with the function  $f(\alpha, \beta) = \mathcal{L}(f(\bm{\theta}^* + \alpha \bm{\delta} + \beta \bm{\eta}; \mathcal{B}^{test}))$, where $\mathcal{B}^{test}$ represents the test set. The resultant contour plots are shown in Fig.~\ref{fig:LOSS_LANDSCAPE}.
As can be seen, the minimum reached by optimising over more AutoML components tend to be flatter than NAS with manual designed DA and HPO.
The loss landscape of DHA is also flatter and smoother than that of Sequential DHA accounting for the better generalisation performance of DHA in previous experiments, explaining the superior test accuracy achieved by DHA in Tables~\ref{table:MAIN_RESULT} and \ref{table:MAIN_RESULT R1}~\citep{keskar2016large, xu2012robustness}.

\subsection{Ablation study of AutoML components}
\label{ablation1}

To further verify the effectiveness of our proposed extended search scope and search strategy, we empirically investigate the performance of considering all three components (i.e., DA, HPO and NAS) against combining any two of them. We examine both the sequential-optimization and joint-optimization settings with four designed optimization algorithms. \textbf{ Sequential NAS + DA} first conducts NAS and during the tuning stage, the proposed DA optimization is applied. Similar to the Sequential NAS+DA, \textbf{Sequential NAS + HPO} firstly conducts NAS. Then with the fixed architecture, Sequential NAS + DA optimizes the HPO and networks' weights simultaneously. In contrast to them, \textbf{Joint-optimization NAS + DA} simultaneously conducts DA strategy optimization, NAS and parameter weights optimization. \textbf{Joint-optimization NAS + HPO} simultaneously conducts HPO, NAS and parameter weights optimization.

The comparison results are presented in Table~\ref{table:ABLATION1}. We can notice that performing joint optimization for either NAS + DA or NAS + HPO, achieves higher test accuracy than doing them in a pipeline. This reconfirms our previous conclusion that optimising different AutoML components jointly is better than doing them in sequence. Moreover, by comparing the results of the joint-optimization NAS+DA and the joint-optimization NAS+HPO in Table~\ref{table:ABLATION1} against the One-Stage ISTA in Table~\ref{table:MAIN_RESULT}, it is clear that considering one more AutoML component on top of NAS can lead to clear performance gain of the final model. While such gain is higher for incorporating DA than HPO, it is maximised when all three components are considered; our DHA obtains the best test accuracy among all joint-optimization baselines across all datasets.

We also provide (1) more ablation studies of DA and HPO (see Appendix~\ref{app:ablation of DA and HPO}), (2) analysis of the searched architectures (see Appendix~\ref{app:analysis of achitectures}) and comparison in terms of FLOPs and Latency (see Appendix~\ref{app:Comparison of Computation Efficiency}). Please refer to the Appendix for more details.

\section{Conclusion}

In this work, we present DHA, an end-to-end joint-optimization method for three important components of AutoML, including DA, HPO and NAS.
This differentiable joint-optimization method can efficiently optimize larger search space than precious AutoML methods and achieve SOTA results on various datasets with a relatively low computational cost. Specifically, DHA achieves 77.4\% Top-1 accuracy on ImageNet with cell based search space, which is higher that current SOTA by 0.5\%. With DHA, we show the advantage of doing joint-optimization of AutoML over doing co-optimization in sequence, and conclude that joint optimization of multiple AutoML components is necessary.

\section{Acknowledgments}

We gratefully acknowledge the support of MindSpore, CANN (Compute Architecture for Neural Networks) and Ascend AI Processor used for this research.

\bibliography{main}
\bibliographystyle{tmlr}

\newpage 

\appendix
\section{Experimental Settings}
\label{app:experimental setting}

\subsection{Search Space}

\textbf{(1) Automated DA.} Following \cite{ho2019population},
we consider 14 different operations for data augmentation, including: \tikz\draw[black,fill=black] (0,0) circle (.5ex); AutoContrast   \tikz\draw[black,fill=black] (0,0) circle (.5ex); Equalize \tikz\draw[black,fill=black] (0,0) circle (.5ex); Rotate   \tikz\draw[black,fill=black] (0,0) circle (.5ex); Posterize \tikz\draw[black,fill=black] (0,0) circle (.5ex); Solarize 
\tikz\draw[black,fill=black] (0,0) circle (.5ex); Color    
\tikz\draw[black,fill=black] (0,0) circle (.5ex); Contrast \tikz\draw[black,fill=black] (0,0) circle (.5ex); Brightness \tikz\draw[black,fill=black] (0,0) circle (.5ex); Sharpness \tikz\draw[black,fill=black] (0,0) circle (.5ex); Shear X  
\tikz\draw[black,fill=black] (0,0) circle (.5ex); Shear Y 
\tikz\draw[black,fill=black] (0,0) circle (.5ex); Translate X \tikz\draw[black,fill=black] (0,0) circle (.5ex); Translate Y \tikz\draw[black,fill=black] (0,0) circle (.5ex); Identity.

The magnitude ranging from 0 to 10 of each operation is randomly sampled from a uniform distribution. At each time, two operations would be sampled according to $\bm{\tau}$ and would be successively applied to each sample.

\noindent \textbf{(2) NAS.} Following \cite{liu2018darts}, we consider the cell-based search space, which regards the whole architecture as a stack of similar cells. Each cell consists of a fixed number of nodes and our model tries to find the best operation combination between different nodes. One difference worth mentioning is that in contrast to DARTS \citep{liu2018darts} which has 8 different operations between two nodes, we adopt the setting in \cite{yang2020ista}, which only considers 7 different operation options between two nodes including: \tikz\draw[black,fill=black] (0,0) circle (.5ex); 3 × 3 Separable Convolutions   \tikz\draw[black,fill=black] (0,0) circle (.5ex); 5 × 5 Separable Convolutions \tikz\draw[black,fill=black] (0,0) circle (.5ex); 3 × 3 Dilated Separable Convolutions   \tikz\draw[black,fill=black] (0,0) circle (.5ex); 5 × 5 Dilated Separable Convolutions  \tikz\draw[black,fill=black] (0,0) circle (.5ex); 3 × 3 Max Pooling 
\tikz\draw[black,fill=black] (0,0) circle (.5ex); 3 × 3 Average Pooling    
\tikz\draw[black,fill=black] (0,0) circle (.5ex); Identity.

To check the generalization of DHA, we also have tested our model with MobilenetV2 search space \citep{tan2019efficientnet}. MobilenetV2 search space consists of Mobilenet blocks with kernel size $\{3,5,7\}$, expansion ratio $\{3,6\}$ and identity operation.

\noindent  \textbf{(3) HPO.} In our model, we consider both the L2 regularization(i.e., weight decay) and the learning rate in the experiments involving the HPO.

\subsection{Setting} 

Experiments are run on 8 NVIDIA V100s under PyTorch-1.3.0 and python3.6. We adopt the hyper-parameter setting in \cite{liu2018darts}. As to the two-stage NAS, for experiments with CIFAR10, CIFAR100, SPORT8, MIT67, and FLOWERS102 during the search phase, the initial channel number and the cell number of the architecture are respectively 16 and 8. During the retraining/tuning phase, the final channel number and the cell number of the architecture are respectively 36 and 20. For experiments with ImageNet, the initial channel number of the architecture is 48 and the cell number of the architecture is 4. During the final retraining/tuning phase, the final channel number and the cell number of the architecture are respectively 48 and 14. As to the one-stage NAS, for experiments with CIFAR10, CIFAR100, SPORT8, MIT67, and FLOWERS102, the channel number and the cell number of the architecture are respectively 36 and 20. For experiments with ImageNet, the channel number and the cell number of the architecture are respectively 48 and 14.

Moreover, the Adam optimizer is used in the DHA optimization process. Learning rate and weight decay are randomly initialized. The main reason for setting the lower bound and upper bound is to prevent negative values and large values. We have set the number of epochs as our stopping criterion. Variances shown in experiment results are mainly caused by setting different random seeds, which are used to show the stability of our model. 

\section{Details of Sequential DHA}
\label{app:sequential DHA}

In the section, we provide detailed descriptions of the sequential DHA (see Algorithm~\ref{alg:sequential}).
Sequential DHA consists of two stages. During the first stage, Sequential DHA performs NAS to find the optimal architecture under certain hyper-parameter settings. In the next stage, Sequential DHA performs the online DA and HPO strategy proposed in our paper and trains the architecture derived from the first stage from scratch. 

\begin{algorithm}[t] 
\caption{Sequential DHA} 
\label{algo:seq_algo}
{\bf Phase I:} \hspace*{0.02in} {\bf Initialization:}
Compressed Representation $\mathbf{b}$, Measurement Matrix $\mathbf{A}$, and Network Parameters $\bm{\theta}$ \\
\hspace*{0.02in} {\bf Input:}
Training Set $\mathcal{D}^{tr}$, Parameters $\mathbf{b}$, $\mathbf{A}$, $\bm{\theta}$, and the iteration number $T$\\
\hspace*{0.02in} {\bf Return:}
$\bm{\alpha}^{\star}$ optimal architecture 
\begin{algorithmic}[1] 
\While{$t<T$}
    \State Separately sample a mini-batch $\mathcal{B}^{tr}$ from $\mathcal{D}^{tr}$;
    \State For each sample $x_k$ in mini-batch $\mathcal{B}^{tr}$;
    \State Recover $\bm{\alpha}^t$ by solving Eqn.(5) with $\mathbf{b}^t$ and $\mathbf{A}$; 
    \State Extract a child network from the super network;
    \State Compute the weighted training loss as Eqn.(2); \label{update_T}
    \State Calculate $\bm{\theta}^{t+1}$ with Eqn.(3);
    \State Use training loss function to update $\mathbf{b}^{t+1}$ through the gradient descent, then $\bm{\alpha}^{t+1}$ is updated with Eqn.(6);\label{update_A}
\EndWhile
\end{algorithmic}

{\bf Phase II:} \hspace*{0.02in} {\bf Initialization:} Data Transformation Parameters $\bm{\tau}$, Hyper-parameters $\bm{\eta}$ and Network Parameters $\bm{\theta}$ \\
\hspace*{0.02in} {\bf Input:}
Training Set $\mathcal{D}^{tr}$, Optimal Architecture Parameters $\bm{\alpha}^{\star}$, Parameters $\bm{\tau}$, $\bm{\eta}$, $\bm{\theta}$, and the iteration number $T$\\
\hspace*{0.02in} {\bf Return:}
$\bm{\theta}$
\begin{algorithmic}[1] 
\While{$t<T$}
    \State Separately sample a mini-batch $\mathcal{B}^{tr}$ from $\mathcal{D}^{tr}$;
    \State For each sample $x_k$ in mini-batch $\mathcal{B}^{tr}$, sample a transformation $\mathcal{T}_k(x_k)$ according to $\bm{\tau}^t$;
    \State Compute the weighted training loss function as Eqn.(2) and update $\bm{\tau}^{t+1}$ accordingly; \label{update_T}
    \State Calculate $\bm{\theta}^{t+1}$ with Eqn.(3);
    \State Compute the training loss function with $\bm{\theta}^{t+1}$ on $\mathcal{D}^{val}$ and update $\bm{\eta}^{t+1}$ with Eqn.(6);\label{update_H}
\EndWhile
\end{algorithmic}
\label{alg:sequential}
\end{algorithm}

\section{Ablation Study of DA and HPO}
\label{app:ablation of DA and HPO}

Besides, we conduct an ablation study on the effect of doing both DA and HPO by comparing sequential DA+HPA and joint-optimization DA+HPO against several previous DA algorithms including one baseline method, Fast Autoaugment (FAA) \citep{lim2019fast}, Population-Based Augment (PBA) \citep{ho2019population}, RandAugment(RA) \citep{cubuk2020randaugment} and AutoHAS \citep{dong2020autohas}. The \textbf{Baseline} method with CIFAR10 and CIFAR100 will standardize the data, use horizontal flips with $50\%$ probability, add zero paddings, implement random crops, and finally apply cutout with $16\times16$ pixels. The \textbf{Sequential DA+HP} first applies DA strategy optimization to find a fixed DA strategy, then re-initializes the parameter weights and tunes them with HPO. The \textbf{Joint-optimization DA+HP} simultaneously conducts DA strategy optimization, HPO, and parameters weights tuning. From Table~\ref{table:ABLATION2}, we can learn two insights. Firstly, compared with previous DA and HPO algorithms, the proposed DA can achieve similar performance with low computational requests. Secondly, the additional consideration of HPO can improve over DA-only approaches and vice versa. Moreover, the joint-optimization of both components performs better than sequential-optimization of them.

\begin{table}[!htbp]
\caption{Top-1 Accuracy ($\%$) comparison of different baseline networks with different DA optimization and HPO strategies on CIFAR10, CIFAR100 and ImageNet with Cell-Based Search Space. Listed algorithms includes Baseline algorithm, FAA, PBA, RA, AutoHAS, Sequential DA+HP and Joint-optimization DA+HP.}
\begin{center}
\scalebox{0.85}{\begin{tabular}{c|c|c|c}
\hline
\hline
Strategy & CIFAR10 & CIFAR100  & ImageNet  \\
\hline
\hline
 &  Wide-ResNet-28-10 &  Wide-ResNet-40-2 &  ResNet-50 \\
\hline
\hline
Baseline & $96.1$ & $74.0$ & $76.3/93.1$\\
FAA & $97.3$ & $79.3$ & $77.6/93.7$\\
PBA & $97.4$ & - & -\\
RA & $97.3$ & - & -\\
\hline
AutoHAS  & -  & - & $78.5/-$ \\
\hline
Sequential DA+HPO & $97.5$ & $79.5$ & - \\
\hline
Joint-optimization DA+HPO  & $\bm{98.0}$  & $\bm{80.6}$ & $\bm{79.0/94.3}$ \\
\hline
\hline
\end{tabular}
}
\label{table:ABLATION2}
\end{center}
\end{table}

\section{Analysis of Searched Architectures}
\label{app:analysis of achitectures}

\begin{figure}[t]
\setlength{\abovecaptionskip}{0.cm}
\setlength{\belowcaptionskip}{-0.2cm}
    \centering
    \includegraphics[width=13cm]{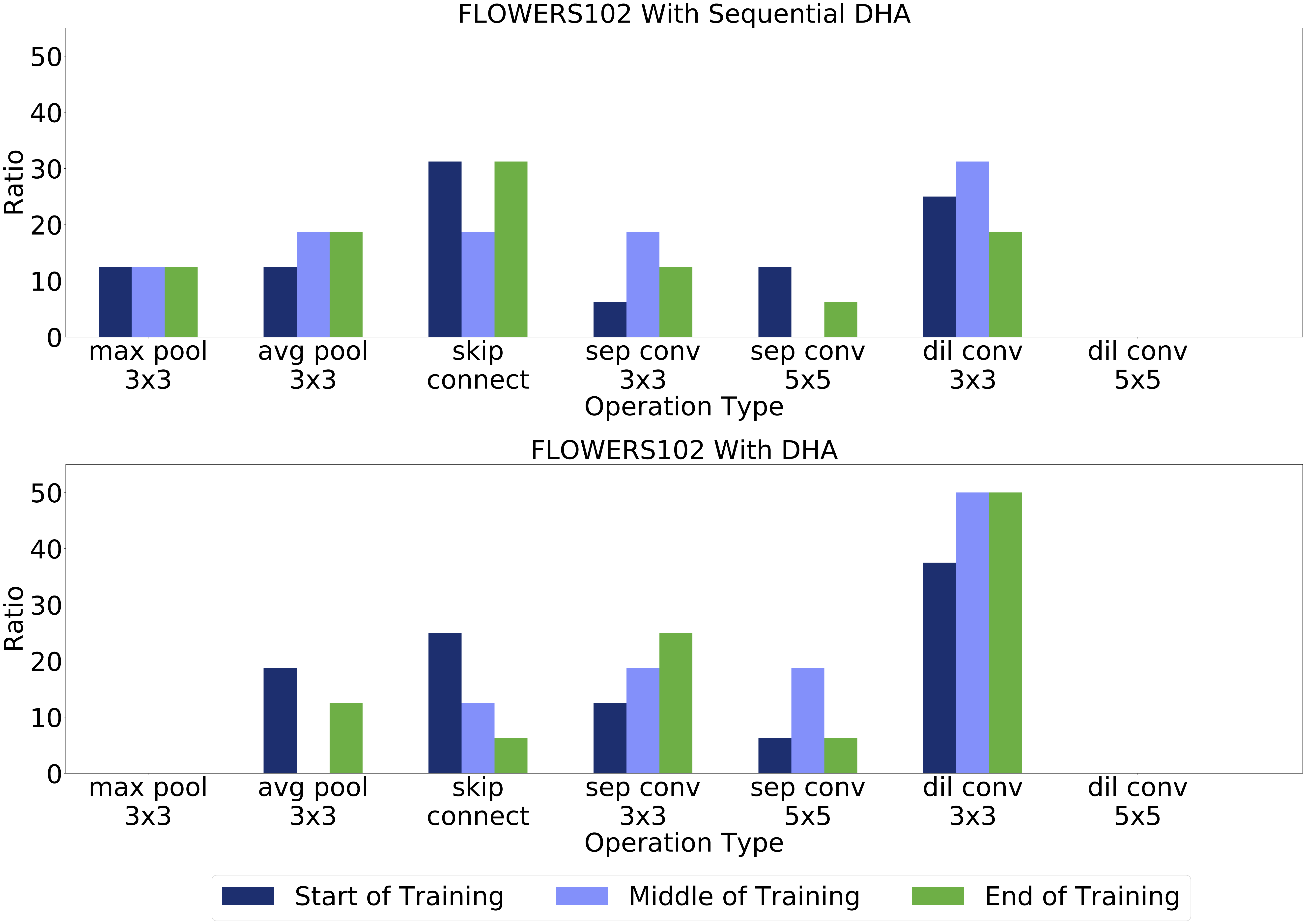}
    \caption{Temporal evolution of Cell-Based search architecture.}
    \label{fig:architecture2}
\end{figure}

\noindent \textbf{Temporal stability of searched architecture.}
The architectures discovered by Sequential DHA and DHA on Flowers102 are shown in Fig.~\ref{fig:architecture2}. Differences between the two found architectures reveal two insights. Firstly, architecture found by Sequential DHA includes almost all kinds of operations and each of them occupies a similar portion. While, for the architecture found by DHA, few operations will occupy relatively large portion. Secondly, for the architecture found by DHA, operation with learn-able weights, e.g., 3 × 3 Dilated Separable Convolutions, will play an important role, while for the architecture found by Sequential DHA, non-weights operation, e.g., Identity will also play an essential actor. With the non-weights operation, during the search stage, it can easily lead to over-fitting, which prevents the algorithm to explore more possible operations. 

\section{Comparison of Computation Efficiency}
\label{app:Comparison of Computation Efficiency}

In this section, we compare the computation efficiency of DHA and the baseline methods in Table~\ref{table:MAIN_RESULT R1}.
As shown in Table~\ref{table:Two_Metrics}, DHA also outperform the baseline methods in terms of FLOPs and latency.

\begin{table*}[hbt!]
\centering
\caption{Comparison in terms of FLOPs and Latency with SOTA image classifiers on ImageNet in the Cell-Based (C) setting or MobileNet/ShuffleNet (M) setting.}
\scalebox{0.82}{
\begin{tabular}{c|c|c|c|c|c|c}
\hline\hline
\textbf{Model} & \multicolumn{2}{c|}{\textbf{Test Acc (\%)}} & \textbf{Params} & \textbf{Flops} & \textbf{Latency} & \textbf{Search}\\ 
\cline{2-3}
 & Top-1 & Top-5 & \textbf{(M)} & \textbf{(M)} & \textbf{(ms)} & \textbf{Space} \\ 
\hline\hline
DARTS (2nd) & 73.3 & 91.3 & 4.7 & 574 & 17.53 & C \\
SNAS (mild) & 72.7 & 90.8 & 4.3 & 522 &  - & C  \\
Two-Stage ISTA & 75.0 & 91.9 & 5.3  & 550 & -  & C \\
One-Stage ISTA & 76.0 & 92.9 & 5.7 & 638 & -  & C \\
\hline
ProxylessNAS (GPU) & 75.1 & 92.5 & 7.1 & 465 & 12.02 & M \\
DSNAS & 74.3 & 91.9 & - & 324 & -  & M \\
APQ & 75.1 & - & - & - & 12.17  & M \\ 
\hline\hline
Sequential DHA & 76.7 & 93.8 & 5.4  & 562 & 16.99  & C \\
DHA & 77.4 & 94.6 & 5.6 & 540 & 17.23 & C\\
\hline
Sequential DHA & 77.1 & 93.9 & 5.7 & 355 & 12.28  & M \\
DHA & \textbf{77.6} & 94.8 & 5.3 & 337 &  12.19 & M \\
\hline\hline
\end{tabular}}
\label{table:Two_Metrics}
\end{table*}

\section{Discussion}
\label{app:discussion}

\noindent \textbf{Discussion about novelty.}
To the best of our knowledge, we are the first that aims to joint-optimize DA, HPO, and NAS. Specifically, instead of training the network under each setting from scratch, we propose to apply weights sharing technology to all three modules. In this way, we can make use of previous training results to evaluate the performance of each setting. Hence, we can deal with huge search space efficiently. Experimental results also prove the effectiveness of DHA on ImageNet and other datasets.


\noindent \textbf{Limitation of DHA.}
Our method mainly concentrates on optimizing the differentiable hyper-parameter like weight decay and momentum. According to \citep{li2020rethinking}, learning rate and weight decay are the two most important hyper-parameters in model training. Experimental results also show the effectiveness of our search space. 

\end{document}

%% file: math_commands.tex
\usepackage{amsmath,amsfonts,bm}

\def\eqref#1{equation~\ref{#1}}

\def\1{\bm{1}}

\DeclareMathAlphabet{\mathsfit}{\encodingdefault}{\sfdefault}{m}{sl}
\SetMathAlphabet{\mathsfit}{bold}{\encodingdefault}{\sfdefault}{bx}{n}













%% file: main.bbl
\begin{thebibliography}{73}
\providecommand{\natexlab}[1]{#1}
\providecommand{\url}[1]{\texttt{#1}}
\expandafter\ifx\csname urlstyle\endcsname\relax
  \providecommand{\doi}[1]{doi: #1}\else
  \providecommand{\doi}{doi: \begingroup \urlstyle{rm}\Url}\fi

\bibitem[Alvi et~al.(2019)Alvi, Ru, Calliess, Roberts, and
  Osborne]{alvi2019asynchronous}
Ahsan~S Alvi, Binxin Ru, Jan Calliess, Stephen~J Roberts, and Michael~A
  Osborne.
\newblock Asynchronous batch bayesian optimisation with improved local
  penalisation.
\newblock In \emph{ICML}, 2019.

\bibitem[Beck \& Teboulle(2009)Beck and Teboulle]{beck2009fast}
Amir Beck and Marc Teboulle.
\newblock A fast iterative shrinkage-thresholding algorithm for linear inverse
  problems.
\newblock \emph{SIAM Journal on Imaging Sciences}, 2\penalty0 (1):\penalty0
  183--202, 2009.

\bibitem[Bengio(2000)]{bengio2000gradient}
Yoshua Bengio.
\newblock Gradient-based optimization of hyperparameters.
\newblock \emph{Neural Computation}, 12\penalty0 (8):\penalty0 1889--1900,
  2000.

\bibitem[Cai et~al.(2019)Cai, Zhu, and Han]{cai2018proxylessnas}
Han Cai, Ligeng Zhu, and Song Han.
\newblock Proxylessnas: Direct neural architecture search on target task and
  hardware.
\newblock In \emph{ICLR}, 2019.

\bibitem[Cai et~al.(2020)Cai, Gan, Wang, Zhang, and Han]{cai2019once}
Han Cai, Chuang Gan, Tianzhe Wang, Zhekai Zhang, and Song Han.
\newblock Once-for-all: Train one network and specialize it for efficient
  deployment.
\newblock In \emph{ICLR}, 2020.

\bibitem[Carlucci et~al.(2019)Carlucci, Esperan{\c{c}}a, Singh, Gabillon, Yang,
  Xu, Chen, and Wang]{carlucci2019manas}
Fabio~Maria Carlucci, Pedro~M Esperan{\c{c}}a, Marco Singh, Victor Gabillon,
  Antoine Yang, Hang Xu, Zewei Chen, and Jun Wang.
\newblock Manas: Multi-agent neural architecture search.
\newblock \emph{arXiv preprint arXiv:1909.01051}, 2019.

\bibitem[Casale et~al.(2019)Casale, Gordon, and Fusi]{casale2019probabilistic}
Francesco~Paolo Casale, Jonathan Gordon, and Nicolo Fusi.
\newblock Probabilistic neural architecture search.
\newblock \emph{arXiv preprint arXiv:1902.05116}, 2019.

\bibitem[Chen et~al.(2021)Chen, Wang, Cheng, Tang, and Hsieh]{chen2020drnas}
Xiangning Chen, Ruochen Wang, Minhao Cheng, Xiaocheng Tang, and Cho-Jui Hsieh.
\newblock Dr{NAS}: Dirichlet neural architecture search.
\newblock In \emph{ICLR}, 2021.

\bibitem[Chen et~al.(2019)Chen, Xie, Wu, and Tian]{chen2019progressive}
Xin Chen, Lingxi Xie, Jun Wu, and Qi~Tian.
\newblock Progressive differentiable architecture search: Bridging the depth
  gap between search and evaluation.
\newblock In \emph{CVPR}, 2019.

\bibitem[Cubuk et~al.(2018)Cubuk, Zoph, Mane, Vasudevan, and
  Le]{cubuk2018autoaugment}
Ekin~D Cubuk, Barret Zoph, Dandelion Mane, Vijay Vasudevan, and Quoc~V Le.
\newblock Autoaugment: Learning augmentation policies from data.
\newblock \emph{arXiv preprint arXiv:1805.09501}, 2018.

\bibitem[Cubuk et~al.(2020)Cubuk, Zoph, Shlens, and Le]{cubuk2020randaugment}
Ekin~D Cubuk, Barret Zoph, Jonathon Shlens, and Quoc~V Le.
\newblock Randaugment: practical automated data augmentation with a reduced
  search space.
\newblock In \emph{CVPR Workshops}, 2020.

\bibitem[Dai et~al.(2020)Dai, Wan, Zhang, Wu, He, Wei, Chen, Tian, Yu, Vajda,
  et~al.]{dai2020fbnetv3}
Xiaoliang Dai, Alvin Wan, Peizhao Zhang, Bichen Wu, Zijian He, Zhen Wei, Kan
  Chen, Yuandong Tian, Matthew Yu, Peter Vajda, et~al.
\newblock {FBNetV3}: Joint architecture-recipe search using neural acquisition
  function.
\newblock In \emph{CVPR}, 2020.

\bibitem[Daubechies et~al.(2004)Daubechies, Defrise, and
  De~Mol]{daubechies2004iterative}
Ingrid Daubechies, Michel Defrise, and Christine De~Mol.
\newblock An iterative thresholding algorithm for linear inverse problems with
  a sparsity constraint.
\newblock \emph{Communications on Pure and Applied Mathematics: A Journal
  Issued by the Courant Institute of Mathematical Sciences}, 57\penalty0
  (11):\penalty0 1413--1457, 2004.

\bibitem[Dong \& Yang(2019)Dong and Yang]{dong2019search}
Xuanyi Dong and Yi~Yang.
\newblock Searching for a robust neural architecture in four gpu hours.
\newblock In \emph{CVPR}, 2019.

\bibitem[Dong et~al.(2020)Dong, Tan, Yu, Peng, Gabrys, and Le]{dong2020autohas}
Xuanyi Dong, Mingxing Tan, Adams~Wei Yu, Daiyi Peng, Bogdan Gabrys, and Quoc~V
  Le.
\newblock {AutoHAS}: Differentiable hyper-parameter and architecture search.
\newblock \emph{arXiv preprint arXiv:2006.03656}, 2020.

\bibitem[Elsken et~al.(2019)Elsken, Metzen, and Hutter]{elsken2018efficient}
Thomas Elsken, Jan~Hendrik Metzen, and Frank Hutter.
\newblock Efficient multi-objective neural architecture search via lamarckian
  evolution.
\newblock In \emph{ICLR}, 2019.

\bibitem[Falkner et~al.(2018)Falkner, Klein, and Hutter]{falkner2018bohb}
Stefan Falkner, Aaron Klein, and Frank Hutter.
\newblock Bohb: Robust and efficient hyperparameter optimization at scale.
\newblock In \emph{ICML}, 2018.

\bibitem[Gonz{\'a}lez et~al.(2016)Gonz{\'a}lez, Dai, Hennig, and
  Lawrence]{Gonzalez_2015Batch}
Javier Gonz{\'a}lez, Zhenwen Dai, Philipp Hennig, and Neil~D Lawrence.
\newblock Batch {B}ayesian optimization via local penalization.
\newblock In \emph{International Conference on Artificial Intelligence and
  Statistics}, 2016.

\bibitem[Guo et~al.(2019)Guo, Zhang, Mu, Heng, Liu, Wei, and
  Sun]{guo2019single}
Zichao Guo, Xiangyu Zhang, Haoyuan Mu, Wen Heng, Zechun Liu, Yichen Wei, and
  Jian Sun.
\newblock Single path one-shot neural architecture search with uniform
  sampling.
\newblock \emph{arXiv preprint arXiv:1904.00420}, 2019.

\bibitem[He et~al.(2016)He, Zhang, Ren, and Sun]{he2016deep}
Kaiming He, Xiangyu Zhang, Shaoqing Ren, and Jian Sun.
\newblock Deep residual learning for image recognition.
\newblock In \emph{CVPR}, 2016.

\bibitem[Ho et~al.(2019)Ho, Liang, Chen, Stoica, and Abbeel]{ho2019population}
Daniel Ho, Eric Liang, Xi~Chen, Ion Stoica, and Pieter Abbeel.
\newblock Population based augmentation: Efficient learning of augmentation
  policy schedules.
\newblock In \emph{ICML}, 2019.

\bibitem[Hu et~al.(2020)Hu, Xie, Zheng, Liu, Shi, Liu, and Lin]{hu2020dsnas}
Shoukang Hu, Sirui Xie, Hehui Zheng, Chunxiao Liu, Jianping Shi, Xunying Liu,
  and Dahua Lin.
\newblock {DSNAS}: Direct neural architecture search without parameter
  retraining.
\newblock In \emph{CVPR}, 2020.

\bibitem[Im et~al.(2021)Im, Savin, and Cho]{im2021online}
Daniel~Jiwoong Im, Cristina Savin, and Kyunghyun Cho.
\newblock Online hyperparameter optimization by real-time recurrent learning.
\newblock \emph{arXiv preprint arXiv:2102.07813}, 2021.

\bibitem[Kandasamy et~al.(2017)Kandasamy, Dasarathy, Schneider, and
  P{\'o}czos]{kandasamy2017multi}
Kirthevasan Kandasamy, Gautam Dasarathy, Jeff Schneider, and Barnab{\'a}s
  P{\'o}czos.
\newblock Multi-fidelity {B}ayesian optimisation with continuous
  approximations.
\newblock In \emph{ICML}, 2017.

\bibitem[Kandasamy et~al.(2018)Kandasamy, Neiswanger, Schneider, Poczos, and
  Xing]{kandasamy2018neural}
Kirthevasan Kandasamy, Willie Neiswanger, Jeff Schneider, Barnabas Poczos, and
  Eric~P Xing.
\newblock Neural architecture search with bayesian optimisation and optimal
  transport.
\newblock In \emph{NeurIPS}, 2018.

\bibitem[Kashima et~al.(2020)Kashima, Yamada, and
  Saito]{DBLP:journals/corr/abs-2012-09407}
Taiga Kashima, Yoshihiro Yamada, and Shunta Saito.
\newblock Joint search of data augmentation policies and network architectures.
\newblock In \emph{CoRR}, 2020.

\bibitem[Kathuria et~al.(2016)Kathuria, Deshpande, and
  Kohli]{Kathuria_NIPS2016Batched}
Tarun Kathuria, Amit Deshpande, and Pushmeet Kohli.
\newblock Batched {G}aussian process bandit optimization via determinantal
  point processes.
\newblock In \emph{NeurIPS}, 2016.

\bibitem[Keskar et~al.(2016)Keskar, Mudigere, Nocedal, Smelyanskiy, and
  Tang]{keskar2016large}
Nitish~Shirish Keskar, Dheevatsa Mudigere, Jorge Nocedal, Mikhail Smelyanskiy,
  and Ping Tak~Peter Tang.
\newblock On large-batch training for deep learning: Generalization gap and
  sharp minima.
\newblock \emph{arXiv preprint arXiv:1609.04836}, 2016.

\bibitem[Krizhevsky et~al.(2009)Krizhevsky, Hinton,
  et~al.]{krizhevsky2009learning}
Alex Krizhevsky, Geoffrey Hinton, et~al.
\newblock Learning multiple layers of features from tiny images.
\newblock \emph{Tech Report}, 2009.

\bibitem[Li et~al.(2017)Li, Xu, Taylor, Studer, and
  Goldstein]{li2017visualizing}
Hao Li, Zheng Xu, Gavin Taylor, Christoph Studer, and Tom Goldstein.
\newblock Visualizing the loss landscape of neural nets.
\newblock \emph{arXiv preprint arXiv:1712.09913}, 2017.

\bibitem[Li et~al.(2020)Li, Chaudhari, Yang, Lam, Ravichandran, Bhotika, and
  Soatto]{li2020rethinking}
Hao Li, Pratik Chaudhari, Hao Yang, Michael Lam, Avinash Ravichandran, Rahul
  Bhotika, and Stefano Soatto.
\newblock Rethinking the hyperparameters for fine-tuning.
\newblock \emph{arXiv preprint arXiv:2002.11770}, 2020.

\bibitem[Li \& Fei-Fei(2007)Li and Fei-Fei]{li2007and}
Li-Jia Li and Li~Fei-Fei.
\newblock What, where and who? classifying events by scene and object
  recognition.
\newblock In \emph{ICCV}, 2007.

\bibitem[Li et~al.(2021)Li, Khodak, Balcan, and Talwalkar]{li2020geometry}
Liam Li, Mikhail Khodak, Maria-Florina Balcan, and Ameet Talwalkar.
\newblock Geometry-aware gradient algorithms for neural architecture search.
\newblock In \emph{ICLR}, 2021.

\bibitem[Lim et~al.(2019)Lim, Kim, Kim, Kim, and Kim]{lim2019fast}
Sungbin Lim, Ildoo Kim, Taesup Kim, Chiheon Kim, and Sungwoong Kim.
\newblock Fast autoaugment.
\newblock In \emph{NeurIPS}, 2019.

\bibitem[Lin et~al.(2019)Lin, Guo, Li, Yuan, Wu, Yan, Lin, and
  Ouyang]{lin2019online}
Chen Lin, Minghao Guo, Chuming Li, Xin Yuan, Wei Wu, Junjie Yan, Dahua Lin, and
  Wanli Ouyang.
\newblock Online hyper-parameter learning for auto-augmentation strategy.
\newblock In \emph{CVPR}, 2019.

\bibitem[Liu et~al.(2019)Liu, Simonyan, and Yang]{liu2018darts}
Hanxiao Liu, Karen Simonyan, and Yiming Yang.
\newblock {DARTS}: Differentiable architecture search.
\newblock In \emph{ICLR}, 2019.

\bibitem[Lorraine et~al.(2020)Lorraine, Vicol, and
  Duvenaud]{lorraine2020optimizing}
Jonathan Lorraine, Paul Vicol, and David Duvenaud.
\newblock Optimizing millions of hyperparameters by implicit differentiation.
\newblock In \emph{International Conference on Artificial Intelligence and
  Statistics}, 2020.

\bibitem[Lu et~al.(2019)Lu, Whalen, Boddeti, Dhebar, Deb, Goodman, and
  Banzhaf]{lu2019nsga}
Zhichao Lu, Ian Whalen, Vishnu Boddeti, Yashesh Dhebar, Kalyanmoy Deb, Erik
  Goodman, and Wolfgang Banzhaf.
\newblock Nsga-net: neural architecture search using multi-objective genetic
  algorithm.
\newblock In \emph{Proceedings of the Genetic and Evolutionary Computation
  Conference}, 2019.

\bibitem[MacKay et~al.(2019)MacKay, Vicol, Lorraine, Duvenaud, and
  Grosse]{mackay2019self}
Matthew MacKay, Paul Vicol, Jon Lorraine, David Duvenaud, and Roger Grosse.
\newblock Self-tuning networks: Bilevel optimization of hyperparameters using
  structured best-response functions.
\newblock \emph{arXiv preprint arXiv:1903.03088}, 2019.

\bibitem[Maclaurin et~al.(2015)Maclaurin, Duvenaud, and
  Adams]{maclaurin2015gradient}
Dougal Maclaurin, David Duvenaud, and Ryan Adams.
\newblock Gradient-based hyperparameter optimization through reversible
  learning.
\newblock In \emph{ICML}, 2015.

\bibitem[Maddison et~al.(2016)Maddison, Mnih, and Teh]{maddison2016concrete}
Chris~J Maddison, Andriy Mnih, and Yee~Whye Teh.
\newblock The concrete distribution: A continuous relaxation of discrete random
  variables.
\newblock \emph{arXiv preprint arXiv:1611.00712}, 2016.

\bibitem[Min et~al.(2020)Min, Gupta, and Ong]{min2020generalizing}
Alan Tan~Wei Min, Abhishek Gupta, and Yew-Soon Ong.
\newblock Generalizing transfer bayesian optimization to source-target
  heterogeneity.
\newblock \emph{IEEE Transactions on Automation Science and Engineering}, 2020.

\bibitem[Mittal et~al.(2020)Mittal, Liu, Karianakis, Fragoso, Chen, and
  Fu]{mittal2020hyperstar}
Gaurav Mittal, Chang Liu, Nikolaos Karianakis, Victor Fragoso, Mei Chen, and
  Yun Fu.
\newblock {HyperSTAR}: Task-aware hyperparameters for deep networks.
\newblock In \emph{CVPR}, 2020.

\bibitem[Nekrasov et~al.(2019)Nekrasov, Chen, Shen, and Reid]{nekrasov2019fast}
Vladimir Nekrasov, Hao Chen, Chunhua Shen, and Ian Reid.
\newblock Fast neural architecture search of compact semantic segmentation
  models via auxiliary cells.
\newblock In \emph{CVPR}, 2019.

\bibitem[Nilsback \& Zisserman(2008)Nilsback and
  Zisserman]{nilsback2008automated}
Maria-Elena Nilsback and Andrew Zisserman.
\newblock Automated flower classification over a large number of classes.
\newblock In \emph{Indian Conference on Computer Vision, Graphics \& Image
  Processing}, 2008.

\bibitem[Perrone et~al.(2018)Perrone, Jenatton, Seeger, and
  Archambeau]{perrone2018scalable}
Valerio Perrone, Rodolphe Jenatton, Matthias Seeger, and C{\'e}dric Archambeau.
\newblock Scalable hyperparameter transfer learning.
\newblock In \emph{International Conference on Neural Information Processing
  Systems}, 2018.

\bibitem[Pham et~al.(2018)Pham, Guan, Zoph, Le, and Dean]{pham2018efficient}
Hieu Pham, Melody Guan, Barret Zoph, Quoc Le, and Jeff Dean.
\newblock Efficient neural architecture search via parameters sharing.
\newblock In \emph{ICML}, 2018.

\bibitem[Quattoni \& Torralba(2009)Quattoni and
  Torralba]{quattoni2009recognizing}
Ariadna Quattoni and Antonio Torralba.
\newblock Recognizing indoor scenes.
\newblock In \emph{CVPR}, 2009.

\bibitem[Real et~al.(2018)Real, Aggarwal, Huang, and Le]{Real2018_AmoebaNet}
Esteban Real, Alok Aggarwal, Yanping Huang, and Quoc~V Le.
\newblock Regularized evolution for image classifier architecture search.
\newblock \emph{arXiv:1802.01548}, 2018.

\bibitem[Ru et~al.(2020{\natexlab{a}})Ru, Alvi, Nguyen, Osborne, and
  Roberts]{ru2019bayesian}
Binxin Ru, Ahsan~S Alvi, Vu~Nguyen, Michael~A Osborne, and Stephen~J Roberts.
\newblock Bayesian optimisation over multiple continuous and categorical
  inputs.
\newblock In \emph{ICML}, 2020{\natexlab{a}}.

\bibitem[Ru et~al.(2020{\natexlab{b}})Ru, Esperanca, and
  Carlucci]{ru2020neural}
Binxin Ru, Pedro Esperanca, and Fabio Carlucci.
\newblock Neural architecture generator optimization.
\newblock In \emph{NeurIPS}, 2020{\natexlab{b}}.

\bibitem[Russakovsky et~al.(2015)Russakovsky, Deng, Su, Krause, Satheesh, Ma,
  Huang, Karpathy, Khosla, Bernstein, et~al.]{russakovsky2015imagenet}
Olga Russakovsky, Jia Deng, Hao Su, Jonathan Krause, Sanjeev Satheesh, Sean Ma,
  Zhiheng Huang, Andrej Karpathy, Aditya Khosla, Michael Bernstein, et~al.
\newblock Imagenet large scale visual recognition challenge.
\newblock \emph{IJCV}, 2015.

\bibitem[Sandler et~al.(2018)Sandler, Howard, Zhu, Zhmoginov, and
  Chen]{sandler2018mobilenetv2}
Mark Sandler, Andrew Howard, Menglong Zhu, Andrey Zhmoginov, and Liang-Chieh
  Chen.
\newblock {MobilenetV2}: Inverted residuals and linear bottlenecks.
\newblock In \emph{CVPR}, 2018.

\bibitem[Shaban et~al.(2019)Shaban, Cheng, Hatch, and
  Boots]{shaban2019truncated}
Amirreza Shaban, Ching-An Cheng, Nathan Hatch, and Byron Boots.
\newblock Truncated back-propagation for bilevel optimization.
\newblock In \emph{International Conference on Artificial Intelligence and
  Statistics}, 2019.

\bibitem[Swersky et~al.(2013)Swersky, Snoek, and Adams]{swersky2013multi}
Kevin Swersky, Jasper Snoek, and Ryan~Prescott Adams.
\newblock Multi-task bayesian optimization.
\newblock In \emph{NeurIPS}, 2013.

\bibitem[Tan \& Le(2019)Tan and Le]{tan2019efficientnet}
Mingxing Tan and Quoc Le.
\newblock Efficient{N}et: Rethinking model scaling for convolutional neural
  networks.
\newblock In \emph{ICML}, 2019.

\bibitem[Tang et~al.(2020)Tang, Gao, Karlinsky, Sattigeri, Feris, and
  Metaxas]{tang2020onlineaugment}
Zhiqiang Tang, Yunhe Gao, Leonid Karlinsky, Prasanna Sattigeri, Rogerio Feris,
  and Dimitris Metaxas.
\newblock Onlineaugment: Online data augmentation with less domain knowledge.
\newblock In \emph{ECCV}, 2020.

\bibitem[Tibshirani(1996)]{tibshirani1996regression}
Robert Tibshirani.
\newblock Regression shrinkage and selection via the lasso.
\newblock \emph{Journal of the Royal Statistical Society: Series B
  (Methodological)}, 58\penalty0 (1):\penalty0 267--288, 1996.

\bibitem[Wang et~al.(2020)Wang, Wang, Cai, Lin, Liu, Wang, Lin, and
  Han]{wang2020apq}
Tianzhe Wang, Kuan Wang, Han Cai, Ji~Lin, Zhijian Liu, Hanrui Wang, Yujun Lin,
  and Song Han.
\newblock {APQ}: Joint search for network architecture, pruning and
  quantization policy.
\newblock In \emph{CVPR}, 2020.

\bibitem[Wei et~al.(2020)Wei, Xiao, Xie, Chen, Zhang, and
  Tian]{DBLP:journals/corr/abs-2003-11342}
Longhui Wei, An~Xiao, Lingxi Xie, Xin Chen, Xiaopeng Zhang, and Qi~Tian.
\newblock Circumventing outliers of autoaugment with knowledge distillation.
\newblock In \emph{CoRR}, 2020.

\bibitem[Williams(1992)]{williams1992simple}
Ronald~J Williams.
\newblock Simple statistical gradient-following algorithms for connectionist
  reinforcement learning.
\newblock \emph{Machine Learning}, 8\penalty0 (3-4):\penalty0 229--256, 1992.

\bibitem[Wu et~al.(2019)Wu, Toscano-Palmerin, Frazier, and
  Wilson]{wu2019practical}
Jian Wu, Saul Toscano-Palmerin, Peter~I Frazier, and Andrew~Gordon Wilson.
\newblock Practical multi-fidelity {B}ayesian optimization for hyperparameter
  tuning.
\newblock In \emph{UAI}, 2019.

\bibitem[Xie et~al.(2019)Xie, Zheng, Liu, and Lin]{xie2018snas}
Sirui Xie, Hehui Zheng, Chunxiao Liu, and Liang Lin.
\newblock {SNAS}: Stochastic neural architecture search.
\newblock In \emph{ICLR}, 2019.

\bibitem[Xu \& Mannor(2012)Xu and Mannor]{xu2012robustness}
Huan Xu and Shie Mannor.
\newblock Robustness and generalization.
\newblock \emph{Machine Learning}, 86\penalty0 (3):\penalty0 391--423, 2012.

\bibitem[Xu et~al.(2020)Xu, Xie, Zhang, Chen, Qi, Tian, and Xiong]{xu2019pc}
Yuhui Xu, Lingxi Xie, Xiaopeng Zhang, Xin Chen, Guo-Jun Qi, Qi~Tian, and
  Hongkai Xiong.
\newblock Pc-darts: Partial channel connections for memory-efficient
  architecture search.
\newblock In \emph{ICLR}, 2020.

\bibitem[Yang et~al.(2019)Yang, Esperan{\c{c}}a, and
  Carlucci]{yang2019evaluation}
Antoine Yang, Pedro~M Esperan{\c{c}}a, and Fabio~M Carlucci.
\newblock Nas evaluation is frustratingly hard.
\newblock \emph{arXiv preprint arXiv:1912.12522}, 2019.

\bibitem[Yang et~al.(2020)Yang, Li, You, Wang, Qian, and Lin]{yang2020ista}
Yibo Yang, Hongyang Li, Shan You, Fei Wang, Chen Qian, and Zhouchen Lin.
\newblock Ista-nas: Efficient and consistent neural architecture search by
  sparse coding.
\newblock \emph{arXiv preprint arXiv:2010.06176}, 2020.

\bibitem[Yao et~al.(2020)Yao, Xu, Zhang, Liang, and Li]{yao2020sm}
Lewei Yao, Hang Xu, Wei Zhang, Xiaodan Liang, and Zhenguo Li.
\newblock Sm-nas: structural-to-modular neural architecture search for object
  detection.
\newblock In \emph{AAAI}, 2020.

\bibitem[Zela et~al.(2018)Zela, Klein, Falkner, and Hutter]{zela2018towards}
Arber Zela, Aaron Klein, Stefan Falkner, and Frank Hutter.
\newblock Towards automated deep learning: Efficient joint neural architecture
  and hyperparameter search.
\newblock \emph{arXiv preprint arXiv:1807.06906}, 2018.

\bibitem[Zhang et~al.(2020)Zhang, Wang, Zhang, and Zhong]{zhang2019adversarial}
Xinyu Zhang, Qiang Wang, Jian Zhang, and Zhao Zhong.
\newblock Adversarial autoaugment.
\newblock \emph{ICLR}, 2020.

\bibitem[Zhou et~al.(2021)Zhou, Li, Xie, Chen, Hong, Sun, and
  Li]{zhou2021metaaugment}
Fengwei Zhou, Jiawei Li, Chuanlong Xie, Fei Chen, Lanqing Hong, Rui Sun, and
  Zhenguo Li.
\newblock Metaaugment: Sample-aware data augmentation policy learning.
\newblock In \emph{AAAI}, 2021.

\bibitem[Zhou et~al.(2019)Zhou, Yang, Wang, and Pan]{zhou2019bayesnas}
Hongpeng Zhou, Minghao Yang, Jun Wang, and Wei Pan.
\newblock Bayesnas: A bayesian approach for neural architecture search.
\newblock In \emph{ICML}, 2019.

\bibitem[Zoph \& Le(2017)Zoph and Le]{ZophLe17_NAS}
Barret Zoph and Quoc Le.
\newblock Neural architecture search with reinforcement learning.
\newblock In \emph{ICLR}, 2017.

\end{thebibliography}
